# Effective Class-Imbalance learning based on SMOTE and Convolutional Neural Networks

**Javad Hasannataj Joloudari[1], Abdolreza Marefat[2], Mohammad Ali Nematollahi[3], Solomon Sunday Oyelere[4], Sadiq Hussain[5]**

[1]Department of Computer Engineering, Faculty of Engineering, University of Birjand, Birjand 9717434765, Iran
[2]Department of Artificial Intelligence, Technical and Engineering Faculty, South Tehran Branch, Islamic Azad University, Tehran, Iran
[3]Department of Computer Sciences, Fasa University, Fasa, Iran
[4]Department of Computer Science, Electrical and Space Engineering, Luleå University of Technology, SE-931 87, Skellefteå, Sweden
[5]Examination Branch, Dibrugarh University, Dibrugarh-786004,Assam, India

**Abstract**
Imbalanced Data (ID) is a problem that deters Machine Learning (ML) models for achieving satisfactory results. ID is the occurrence of a situation where the quantity of the samples belonging to one class outnumbers that of the other by a wide margin, making such models' learning process biased towards the majority class. In recent years, to address this issue, several solutions have been put forward, which opt for either synthetically generating new data for the minority class or reducing the number of majority classes for balancing the data. Hence, in this paper, we investigate the effectiveness of methods based on Deep Neural Networks (DNNs) and Convolutional Neural Networks (CNNs), mixed with a variety of well-known imbalanced data solutions meaning oversampling and undersampling. To evaluate our methods, we have used KEEL, breast cancer, and Z-Alizadeh Sani datasets. In order to achieve reliable results, we conducted our experiments 100 times with randomly shuffled data distributions. The classification results demonstrate that the mixed Synthetic Minority Oversampling Technique (SMOTE)-Normalization-CNN outperforms different methodologies achieving 99.08% accuracy on the 24 imbalanced datasets. Therefore, the proposed mixed model can be applied to imbalanced binary classification problems on other real datasets.

1. Introduction

Learning a classifier from an imbalanced dataset is an important topic and still a complicated problem in supervised learning algorithms. In other words, the class imbalance is a customary long-standing challenge in classification problems (1-5), which deals with a dataset that contains an asymmetrically larger number of samples of the majority class. The imbalanced datasets appear in vast real-world research, such as life sciences (6), facial age approximation (7), anomaly detection (8), determining counterfeit credit card transactions (9), medical imaging (10), DNA sequence identification (11) and so forth. For an imbalanced binary classification problem, samples are typically characterized by two classes namely majority and minority.

In general terms, the minority class often illustrates samples of higher importance and interest rather than the majority class. Nevertheless, compared to the minority class, the majority class usually has a more significant number of samples in a meaningful way, and sometimes, the situation may be extremely serious.

Different situations can occur in confronting the imbalanced datasets, and four common cases are depicted in Figure 1, where the blue-filled circles represent the samples of the majority class, in contrast, the red circles denote the minority class (12). It has been shown that the type of data complexity is the principal determining factor of classification performance reduction (13).

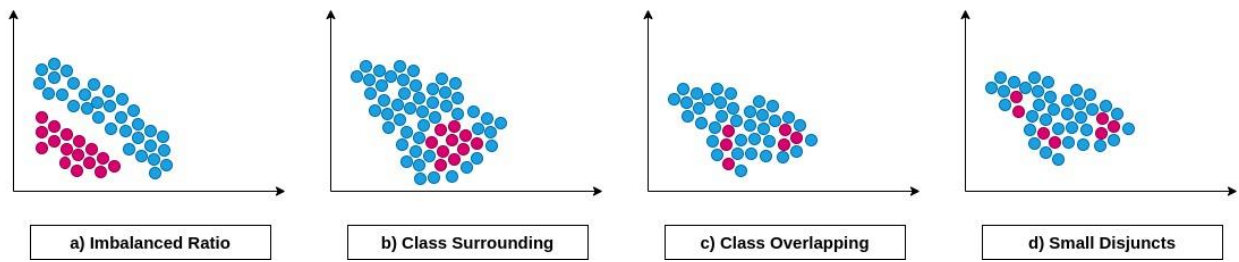

Fig. 1. The schematic diagram of several data distributions with two-dimensional binary-class-imbalanced data.

Most of the classical classification methods, like decision trees (13-15), KNN (16, 17), and Repeated Incremental Pruning to Produce Error Reduction (RIPPER) (18, 19), usually train models that maximize the accuracy of proposed algorithms, sometimes ignoring the minority class (20-22). Hence, several techniques have been designed and implemented to handle the imbalanced binary classification problems. Among these techniques, oversampling and undersampling are well-known (1, 23-25). Yet, the common undersampling and oversampling algorithms modify the initial class distribution of the dataset by excluding the majority class samples or expanding the minority class samples. Cost-sensitive learning algorithms were among the solutions for the above-mentioned issues of imbalanced data (26-28). Such algorithms designate misclassification cost errors for multiple classes, mainly lower costs for the samples of the majority class and higher for the minority class. In addition, Bagging (29) and Boosting (30) methods, which are based on ensemble learning algorithms, are among the other commonly used methods to handle imbalanced class problems (18, 19, 31, 32)

In this paper, we use several undersampling and oversampling methods in the process of implementing our methodology, which is briefly introduced in the sequel:

1) RUS: Among undersampling methods, random undersampling (RUS) is the simplest one, in which the samples of the majority class are randomly removed until suitable balanced data is obtained (33).
2) Tomek Links: Some of the undersampling techniques focus on overlap elimination. For example, the Tomek Links (34) method, which.) is a modification of the Condensed Nearest Neighbor rule, is one of these methods.
3) One-Sided Selection: As the development of the Tomek Links algorithm, one can refer to the One-Sided Selection or briefly OSS method (35) that merges Tomek Links and the Condensed Nearest Neighbor algorithms.
4) Near Miss is another popular undersampling method that randomly removes the majority of class samples. When two samples classified in different classes are very close to each other, it removes the sample belonging to the larger class (16).
5) ROS: Among the oversampling algorithms, random oversampling (ROS) is the simplest one that merely selects and copies the samples from the minority class randomly, leading to more balanced data (36).
6) SMOTE: The best-known oversampling method is the Synthetic Minority Oversampling Technique (SMOTE) (24, 37, 38) which leverages the kNN algorithm to identify the neighbors of minority class samples and generates the new sample by selecting the $k^{th}$ neighbor randomly (39)

It is worth noting that the methods mentioned above may cause some unexpected issues. For example, undersampling techniques may ignore some valuable data, which could be vital for training a classifier. In contrast, oversampling algorithms may cause overfitting. Also, for cost-sensitive learning techniques, it is not straightforward to determine the exact misclassification cost, and different misclassification costs might result in different induced outcomes. Moreover, Bagging and Boosting algorithms may exclude some valuable data while they propose sampling methods in every single iteration and they may face an overfitting problem. Consequently, the classification results obtained by these methods are not stable.

To address these problems, this paper proposes two DL-based methods mixed with different resampling methods for better tackling the issue of an imbalanced dataset. The existing DL-based methods, especially CNN architectures, have been employed in a wide variety of challenges, and they have proven to be extremely powerful in terms of learning balanced datasets. Their efficacy has not been satisfactorily investigated when tackling imbalanced datasets (40).

The CNNs are types of architectures that contain convolutional blocks and can provide an end-to-end classification algorithm. These blocks are a stack of different layers, namely convolutional layers, pooling layers, and activation functions. The most significant attributes of such models are their learning capacity with fewer parameters and translational invariance concerning the input data. In CNNs, the input data are fed to multiple convolutional blocks, which are named mainly backbone as a whole, and then followed by a sequence of fully connected layers to be classified.

The training procedure is done using Focal Loss (FL), which optimizes the abstraction learned by the models to handle complex samples better.

In particular, the main contributions of this paper are threefold. First, 24 popular imbalanced datasets from KEEL Dataset Repository, breast cancer dataset from KDD Cup, and Z-Alizadeh Sani are chosen. The proposed pipeline is trained and validated 100 times with the object of achieving more reliable results. Second, this paper is the first research work that extensively investigates the most efficient mix of Deep Neural Networks (DNN) with the famous resampling method, SMOTE, for imbalanced data. Lastly, the mixed SMOTE-Normalization-CNN methodology has proved to produce superior results in comparison with related research works in terms of accuracy, precision, recall, Geometric mean (G-Mean), specificity, Area Under the Receiver Operating Characteristic Curve (AUC-ROC), and Kappa.

The rest of the paper is outlined as follows: Section 2 provides a brief review of the existing works on imbalanced datasets; Section 3 presents the details of our proposed methodology; Section 4 includes the implementation setup and the evaluation process and then focuses on the results obtained by the proposed methods; Section 5 presents a discussion on comparing our method to the others, and the last section draws a reasoned conclusion and future work.

## 2. Related work

To better handle imbalanced data, Li et al. (41) proposed a developed AdaBoost algorithm, called Adaboost-A, which is based on the AUC evaluation metric. In fact, by considering the impact of misclassification probabilities and AUC, the algorithm mentioned above betters the computational performance of the Adaboost algorithm. Also, this study proposed an ensemble learning algorithm named PSOPD-AdaBoost-A, by which the multipliers of Adaboost weak classifiers were optimized.

In (42), the authors provide a detailed exploratory comparison between the problem of handling class overlapping and class imbalances using a full range of class overlap along with a large scale of class imbalance degrees. The rest of this study contains a thorough review of the current methods and solutions for handling the imbalanced data classification problem, characterized by two categories: distribution-based and class overlap-based algorithms.

The literature (43) proposed an undersampling technique that mainly uses the well-known Naïve Bayes classifier. Based on a random primary selection, this classifier is leveraged to select the most informational samples among the existing training dataset. At first, the model is trained on a small training set, and after that, by an iterative teaching method over the current samples, the base model is taught. The practical outcomes showed that the proposed undersampling technique is comparable to other resampling techniques.

In (44), Dablain et al. proposed a novel deep oversampling method, called DeepSMOTE, which contains three principle parts and mainly uses the properties of the effective SMOTE method. Despite being simple, the method is efficient and powerful. An encoder/decoder structure, being SMOTE-based and including an improved loss function, form the three parts of this method. The results of this study show the advantages of the proposed method, especially in GAN-based oversampling cases.

The research study (45) investigates the effects of resampling on the performances of multi-class Artificial Neural Networks. Different resampling techniques (both undersampling and oversampling) were examined on several cyber security datasets. Also, to determine the results of the proposed methods, various evaluation metrics were used. Finally, four observed patterns were reported that compare the impacts of resampling on the evaluation metrics and model training duration.

In another study (46), the authors dealt with the issue of the probable bias and tendency of a learning classifier toward the majority samples for an imbalanced dataset. This work implemented an innovative three-dimensional framework that includes a discriminator, a generator, and a classifier, together with decision boundary regularization. The remarkable aspect of the proposed method is training a generator in association with a classifier. The reporting results show better performance of the technique than the existing methods.

To improve the efficiency and functioning of undersampling methods for imbalanced data, Xie et al. (47) proposed a new undersampling technique that leverages consecutive density peaks to gradually take out samples from the majority class of an imbalanced data. To determine the importance of the samples of the majority class, two factors were considered, which generate a sequence of samples for learning classifiers. The study compared the implemented algorithm to six well-known undersampling methods over 40 public benchmarks, and the results verify the outperformance of the proposed technique.

In order to design learning classifiers that provide stable performances on imbalanced datasets, the study (48) proposed three different methods. These methods are mainly based on genetic algorithms which automatically specify the ratios of samples for oversampling, undersampling, and hybrid sampling techniques. The implemented algorithms were examined on 14 imbalanced datasets, and the results show that they achieved the best AUC compared to random sampling methods.

To decrease the domination of the majority class samples, the study (49) implemented a novel hybrid method called CDSMOTE, which uses class decomposition and oversampling on the minority class samples. Contrary to general undersampling algorithms, this proposed method keeps the majority class samples, leading to more balanced data. The algorithm was examined on 60 imbalanced public datasets, and the results show comparable performance compared to the existing algorithms.

By introducing a new algorithm called SMOTE-LOF, Maulidevi & Surendro (50), attempted to refine SMOTE. This method distinguishes the noise that arises when dealing with imbalanced datasets by adding the Local Outlier Factor (LOF). By examining the proposed algorithm over different imbalanced datasets, the results were compared to SMOTE. Unlike small data samples, for a large-scale dataset with a small imbalance ratio, SMOTE-LOF outperforms the SMOTE.

In order to annihilate the overlap between the majority class and the minority class in an imbalanced dataset and obtain a balanced and normalized class distribution, the study (39) implemented two innovative density-based methods. These methods were density-based undersampling (DB_US) and density-based hybrid sampling (DB_HS). The first method applies merely an undersampling algorithm, while the second implements both undersampling and oversampling approaches. In addition, the balanced datasets were modeled employing Random Forest (RF) and Support Vector Machine (SVM) classifiers. As a result, the two proposed methods eliminated high-density samples from the majority class and omitted the noises of both classes. The performance of these methods was examined on 16 imbalanced datasets.

In the literature (51), the authors proposed a novel classification method called the Bagging Supervised Autoencoder Classifier (BSAC) to model the credit scoring problems. This algorithm essentially leverages the superlative implementation of a supervised autoencoder based on the axioms of multi-task learning. Also, BSAC tackles the issue of imbalanced datasets via engaging a variation of the Bagging procedure based on undersampling techniques. The examinations of benchmark and real-world credit scoring datasets show the robustness and efficiency of BSAC.

To improve the performance of the basic antlion optimization (ALO), in (52), a novel modified antlion optimization method (MALO) was introduced. This algorithm adds an extra variable that depends on the step size of the ants as revising the antlion position. Also, MALO is modified to the issues of sample reduction to achieve better performance due to various metrics. MALO was examined on several benchmarks and balanced and imbalanced datasets. The results show the outperformance of MALO against the primary ALO method and some other comparable algorithms.

Yang et al. in (53) implemented a sampling level technique called gravitational balanced multiple kernel learning (GBMKL) algorithm, which merges the gravity approach to produce the gravitation balanced midpoint samples (GBMS) placed on the classification boundary. Moreover, to better the generalization efficiency, the classification boundary was modified according to the nearest neighbors of the boundary (NNB) samples. Finally, two regularization terms that correspond to GBMS and NNB were formulated to prevent overfitting. The resulting method was examined on 54 artificial and real-life imbalanced datasets, and the outcomes show the dominance of the implemented method.

Tanimoto et al. in (54) studied the near-miss positive samples in the class of imbalanced datasets. They showed that if the true positive samples are severely limited, the accuracy of the proposed model could be increased by obtaining modified label-like side information positivity to identify near-miss samples from true negatives. Also, the proposed method is following learning using privileged information that leverages side information for training the desired model devoid of predicting the side information itself. The results of the experiments show the outperformance of the method in contrast to the existing algorithms.

The research study (55) proposed new development of SMOTE by merging it with the Kalman filter. After applying SMOTE to the given dataset, the implemented algorithm, called Kalman-SMOTE (KSMOTE), excludes the noisy samples in the resultant dataset that simultaneously contains the initial data and the synthetically added samples. The method was examined on a broad range of datasets, and the results show that the implemented algorithm outperforms the existing methods.

Since oversampling techniques cannot usually achieve high performance in the presence of noise, the study (56) implemented an innovative oversampling algorithm, called IR-SMOTE, that handles this issue. By sorting the majority class samples and the k-means clustering algorithm, the noise in minority class clusters is eliminated. After that, using the kernel density estimation method, the amount of synthetic samples is compatibly designated to each cluster. Finally, regarding random-SMOTE, the desired algorithm was improved to add new samples with an ensured diversity.

The literature (40) studied the performance of convolutional neural networks (CNNs) in the presence of imbalanced data for classification problems. To explore this probable impact, the research used MNIST, CIFAR-10, and ImageNet as benchmarks, alongside undersampling, oversampling, two-phase training, and thresholding. The results show that imbalanced data has a detrimental effect on the performance of the proposed method. Also, one should implement oversampling to the level that removes the imbalance, while the extent of imbalance determines the ideal undersampling ratio. In addition, oversampling does not lead to the overfitting of CNNs.

Fault diagnosis of complex equipment, which plays an important role in the industries, is a crucial technology, and CNN is a general tool for this purpose. In this case, faults are not common, which leads to imbalanced data, and therefore, one cannot propose CNN methods directly. To address this problem, a hierarchical training-CNN is implemented in (57). At first, the method uses a number-resampling technique to balance data. Then, a magnet-loss pretraining algorithm is provided to handle the overlap between diverse faults. The proposed method was examined on the public dataset CWRU with an accuracy of 94.28%.

## 3. Methodology

In this paper, we have used our methods applied to various datasets collected from benchmark repositories such as the KEEL[1], breast cancer[2], and Z-Alizadeh Sani[3] datasets in order to address the class imbalance problem. Figure 2 demonstrates an overview of our proposed methodology, whose details are included in this section.

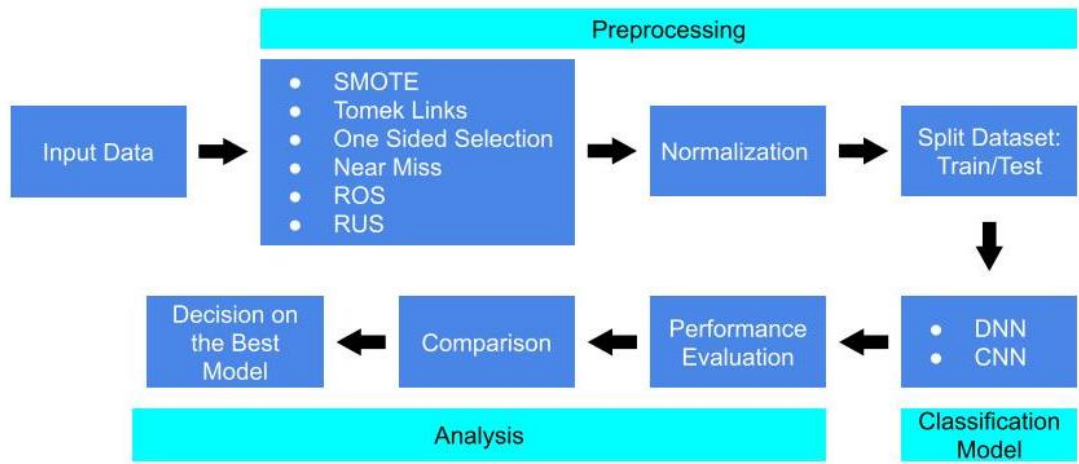

Fig. 2. An overview of our proposed methodology.

Based on Figure 2, the main steps in our methodology include preprocessing, classification, and analysis of models.

### 3.1. Dataset Preprocessing

As stated before, the most acute problem in classifying imbalanced data is that classifiers become biased towards the majority class. There are several methods to overcome this issue which are generally called resampling techniques. By adding minority class samples or removing samples from the majority class, resampling turns the data into a more balanced one. In this regard, there are two principal methods: oversampling and undersampling.

Oversampling algorithms generate new samples, duplicated or synthetic, that belong to the minority class. In contrast, the undersampling techniques delete samples that belong to the majority class to afford balance to the dataset (33).

As a preprocessing step in our methodology, we have utilized various well-known oversampling and undersampling techniques for balancing the dataset. Normalization and split dataset are the next steps in data preprocessing. These are elaborated in the following.

#### 3.1.1. Oversampling techniques

**1) Random Over-Sampling (ROS)**

The first and simplest method in this field is random oversampling (ROS), which aims to help the distribution of datasets by increasing the number of samples in the majority class until the class distributions tend to a balance. This approach is non-heuristic, meaning that it does not boast any intelligent decision boundaries. Random oversampling is usually applied to the level that excludes the imbalance. By merely regenerating samples from the minority class, ROS tackles a balancing in the training model. However, duplicating similar samples may lead to the problem of overfitting, particularly for the samples belonging to the minority classes (36, 58). Figure 3 shows an illustration of the oversampling technique.

---

[1] https://sci2s.ugr.es/keel/imbalanced.php
[2] https://www.kdd.org/kdd-cup
[3] https://archive.ics.uci.edu/ml/datasets/Z-Alizadeh+Sani

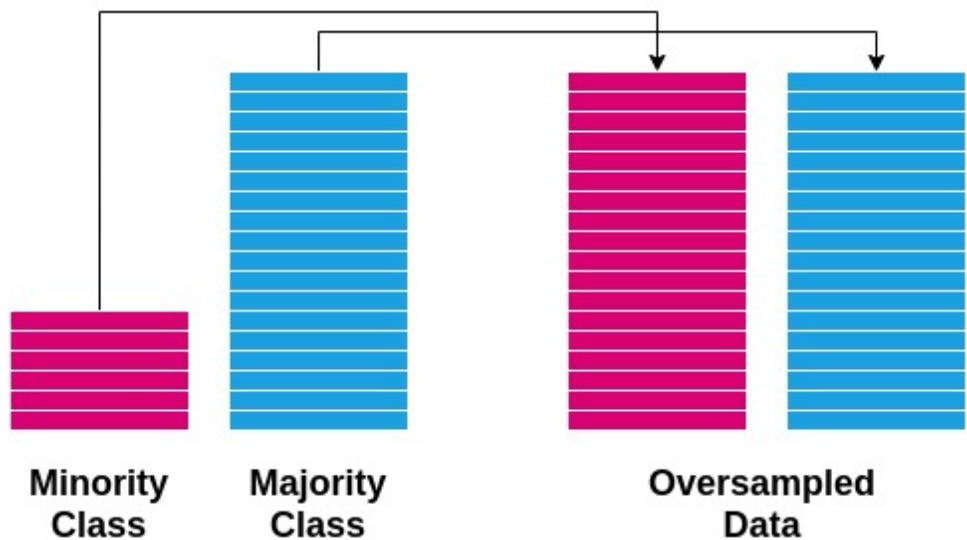

Fig.3. Illustration of Random Over-Sampling.

**2) Synthetic Minority Oversampling Technique**

Synthetic Minority Oversampling Technique (SMOTE) (24, 59-61) is another resampling technique that aims to increase the amount of minority class samples by creating synthetic samples in the minority class and is applied for balancing datasets with a highly unbalanced ratio. In order to avoid the issue of overfitting, the synthetic generation of new samples differed from the multiplication algorithm.

The main idea behind SMOTE is to generate new samples of data in the minority class by interpolation between samples of this class that are in close vicinity of each other (15, 62). Thus, SMOTE increases the number of minority class examples within an imbalanced dataset and consequently enables the classifier to achieve better generalizability. The formal procedure for SMOTE can be explained as follows: Firstly, $N$, which is the desired amount of oversampling, should be set to an integer number. This number can be opted for in that the dataset becomes balanced with a ratio of 1:1 within the different classes. Then, three main steps should be taken iteratively. These steps are 1: Randomly selecting a sample that belongs to the minority class, 2: The $K$ (default 5) nearest neighbors of this sample should be selected, 3: $N$ of these $K$ neighbors are selected randomly for interpolation and generating new samples (63). An intuition of how SMOTE works is shown in Figure 4.

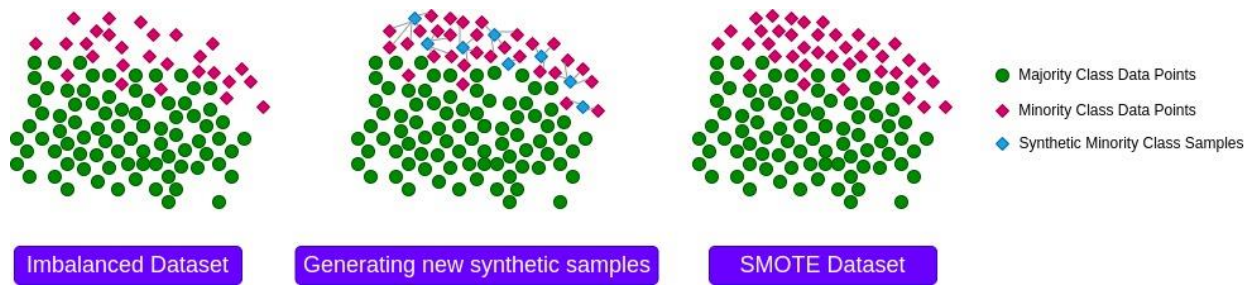

Fig. 4. Illustration of SMOTE Oversampling for Imbalanced Classification.

**3.1.2. Undersampling techniques**

**1) Random Under-Sampling**

The simplest technique among under-sampling methods is Random Under-Sampling (RUS) which is a data-level approach. Here, the algorithm tries to reduce the number of the majority class samples to balance data. In RUS, we randomly select samples within the majority class and delete them, which makes the distribution of a class-imbalanced dataset with a highly unbalanced ratio more balanced. RUS is a non-heuristic approach that does not behave as smart as some other algorithms. Its main drawback is the high probability of losing valuable information within a dataset (15). More precisely, the principal issue in proposing this method is that there is no control over what information about the majority class is being thrown away. As a result, the samples that contain information and details about the decision boundary may be removed, and that valuable information is lost (33). An overview of RUS is shown in Figure 5.

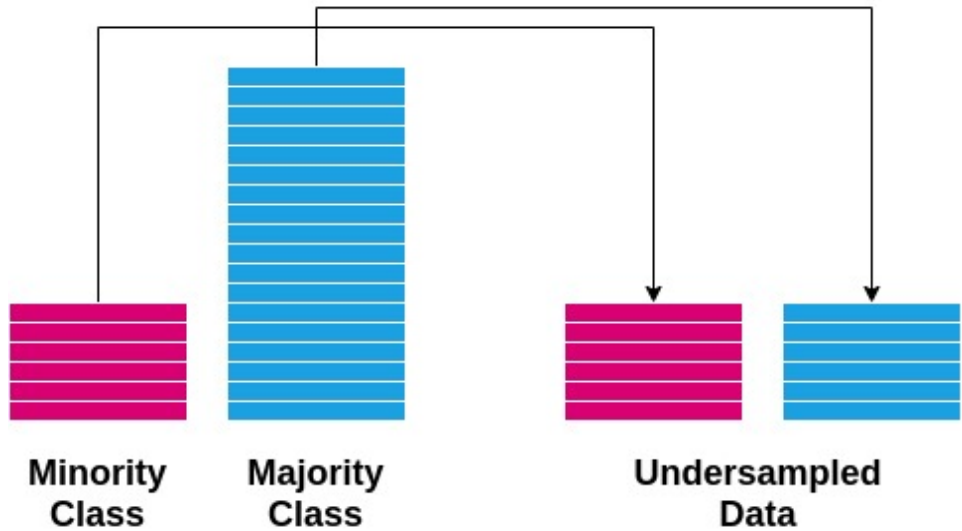

Fig. 5. Illustration of Random Under-Sampling.

**2) Tomek Links**

Tomek Links (TL) (34) is another effective undersampling technique used for balancing the data. TLs are pairs of samples that are very close two each other, but they belong to different classes. These samples are contiguous to the borderline between classes. In mathematical language, given a pair of samples $(S_i. S_j)$ from the dataset, we say that there is a TL between the two samples if at least one of the two following inequalities is satisfied:

$$\delta(S_i. S_l) < \delta(S_i. S_j) \quad , \quad \delta(S_j. S_l) < \delta(S_i. S_j) \quad (1)$$

Where $\delta(x. y)$ is the distance between $x$ and $y$ (64).

Generally, one of the two samples that form a TL is considered a noisy sample, or the two samples together are considered borderline (15). In this case, by eliminating the samples of the majority class that belong to the pairs forming TLs, the distance between the two classes increases, and the dataset becomes more balanced (33). See Figure 6, which shows how TLs can be used to reduce the number of samples in the majority class.

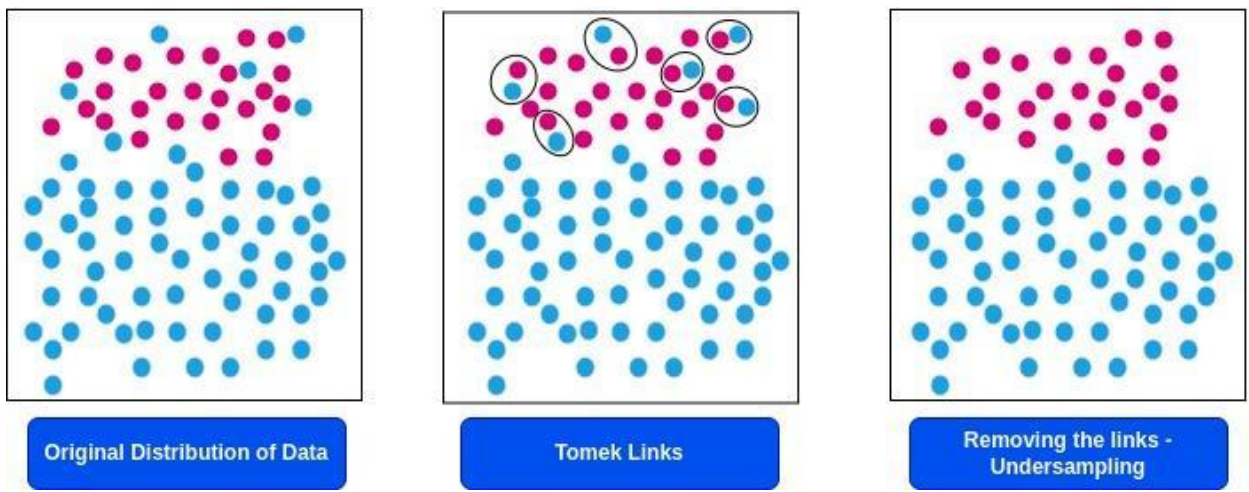

Fig. 6. Illustration of Tomek Links.

### 3) One-Sided Selection (OSS)

One-Sided Selection (OSS) (35) is proposed as an undersampling technique whose main idea is to combine TL and Condensed Nearest Neighbor Rule. To address the issue of imbalanced datasets, this approach leaves the minority class samples completely intact. It filters out the redundant samples in the majority class through a modification of the condensed nearest neighbor rule (65).

In OSS, $\delta(x.y)$ is supposed to be a distance value that meets the requirements for being a TL, where $x$ is chosen from the majority class, and $y$ is selected from the minority. This way, two scenarios can happen: 1) a TL is found to be on the class boundary when both $x$ and $y$ exist in the right class regions 2) a TL is found to be inside one of the class regions when either $x$ or $y$ lies in the wrong region. OSS was introduced to decrease the number of majority class samples by omitting the data points which are borderline or noisy (66). Figure 7 illustrates a diagram of the OSS technique.

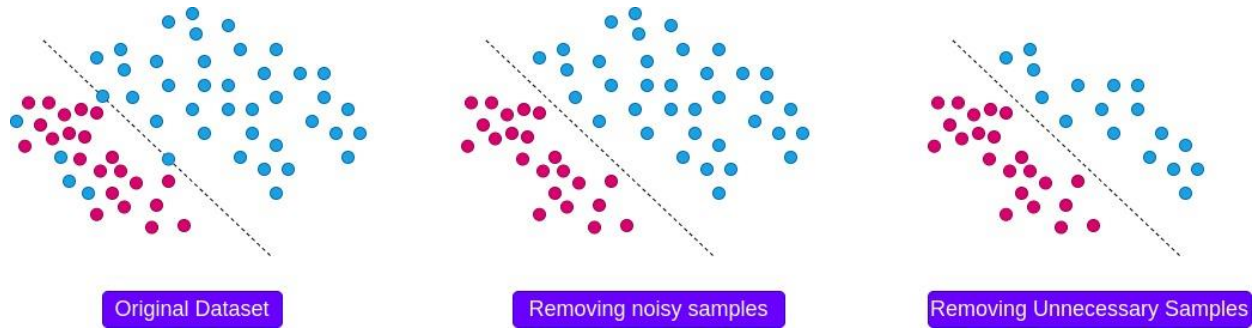

Fig. 7. Illustration of One-Sided Selection.

### 4) NearMiss

The last undersampling technique proposed in this study and introduced here is NearMiss (16). This method is based on the K-nearest neighbors algorithm and categorized as NearMiss-1, NearMiss-2, and Nearmiss-3. The main idea behind NearMiss is to consider the mean distances of samples from the majority class to the samples from the minority class.

Contrary to randomly removing samples from the majority class, these methods eliminate these samples intelligibly. NearMiss-1 removes the majority class samples whose mean distances to the three nearest samples of the minority class are minimal. On the other hand, NearMiss-2 deletes the samples from the majority class with the minimal average distances to the three farthest minority samples. Finally, NearMiss-3 selects a certain number of the closest samples of the majority class regarding every minority class sample (67).

As claimed in (16), the results of experiments showed that NearMiss-2 has a better performance than NearMiss-1 and NearMiss-3. Also, it outperforms the RUS technique (33).

It is worth noting that NearMiss can be fine-tuned in two aspects: The variant that can be chosen from 1, 2, and 3. In addition, the number of neighbors to consider for calculating the mean distances is three as the default. An outline of the NearMiss algorithms is shown in Figure 8.

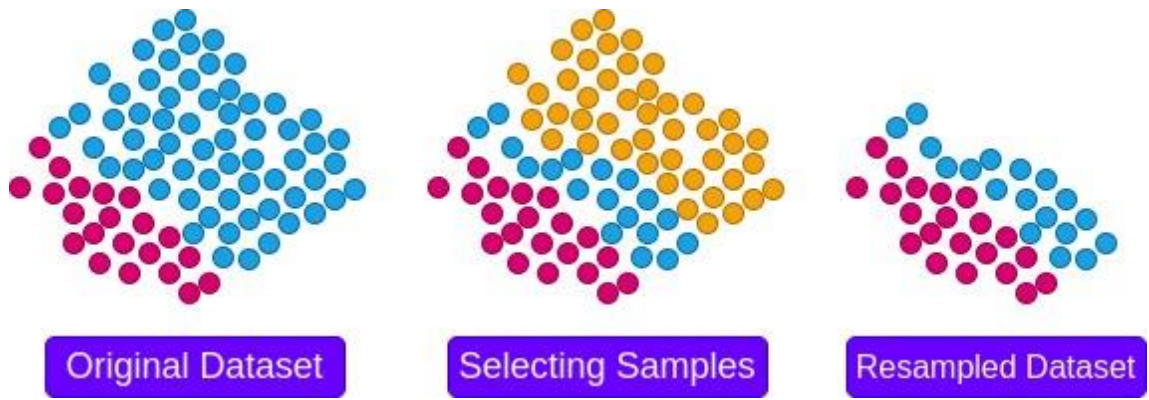

Fig. 8. Illustration of NearMiss.

### 3.1.2. Normalization

Normalization is one of the most crucial preprocessing steps for any challenge in machine learning. It can be done by scaling or transforming the original data to balance the contributions of different features in data samples. In this study, we have normalized the input data to make a distribution between zero and one.

### 3.1.3. Split dataset

Further, due to the low number of samples in datasets which makes the classification result extremely unstable, we have trained and evaluated our models for 100 runs. In each run, first, we randomly shuffle and split the data into training and testing sets and train them according to the model for 2000 epochs and then evaluate it.

## 3.2. Models

In this section, we introduce our two proposed Artificial Neural Network (ANN)-based models, including Deep Neural Network and Convolutional Neural Network.

### 3.2.1. Proposed Deep Neural Network

Deep Neural Networks (DNNs) have recently become among the favorite approaches in various fields in the domain of Artificial Intelligence (68). These networks, which are famously called models, are characterized by several layers that contain a huge number of computational units. These units, which are interconnected, meaning that the output of a unit is the input of the other, are conceived as the imitation of the physiological brain's structure. In mathematical terms, they are a set of parametrized linear and nonlinear transformations capable of being adjusted in order to output abstractions of the input data (69). This capability comes from the amalgamation of multiple layers full of perceptrons. Although a single perceptron cannot handle data that are not linearly separable, they are the basis of Multi-Layer Perceptrons (MLPs), whose ability to transform highly non-linear data makes them a powerful and efficient tool in machine learning (70).

Furthermore, the first proposed method in this paper is a DNN-based model. The architecture of this model is demonstrated in Figure 9.

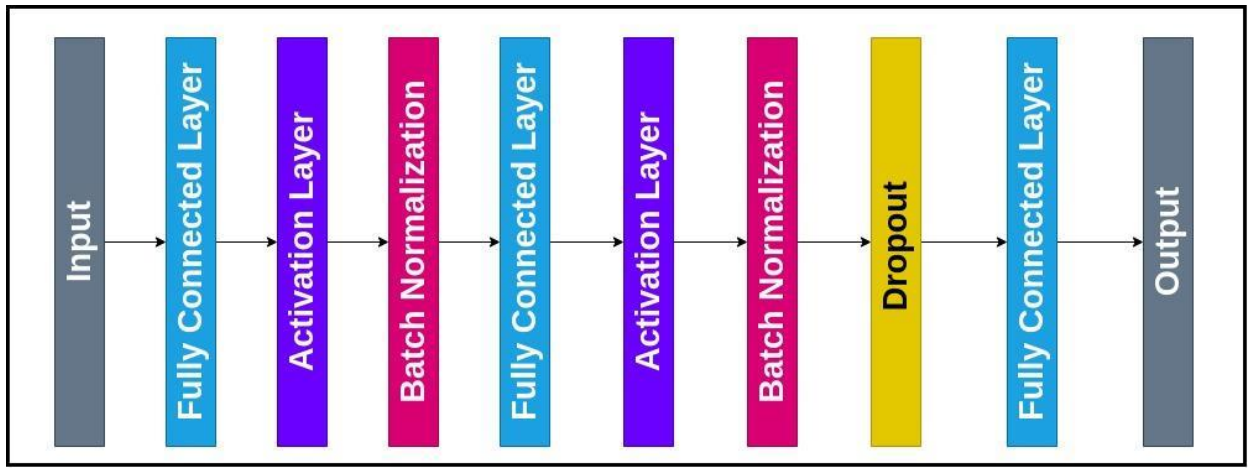

Fig. 9. The architecture of the proposed DNN-based classifier.

As is observed in Figure 9, our DNN model comprises different layers, including a fully connected one followed by an activation function and a batch normalization layer. Then, a fully connected layer, an activation function, and a batch normalization followed by a dropout and a single neuron fully connected layer come after.

### 3.2.2. Proposed Convolutional Neural Network

Convolutional operations are the main components in Convolutional Neural Network (CNN)-based models. These operations enable CNNs to extract and learn the salient features existent in the input data (71). CNN comprises different layers that output feature maps, resulting in sliding different kernels on the input and applying activation functions (72). Compared with DNNs, the major advantage of CNNs over DNNs is their capability to reduce the computational cost in each layer. The convoluted features extracted by these models are compact representations of the input data, which can be further used in downstream tasks such as classification (73).

In this paper, the second method proposed for the binary classification of the input data is a CNN-based model. The architecture of this model is demonstrated in Figure 10.

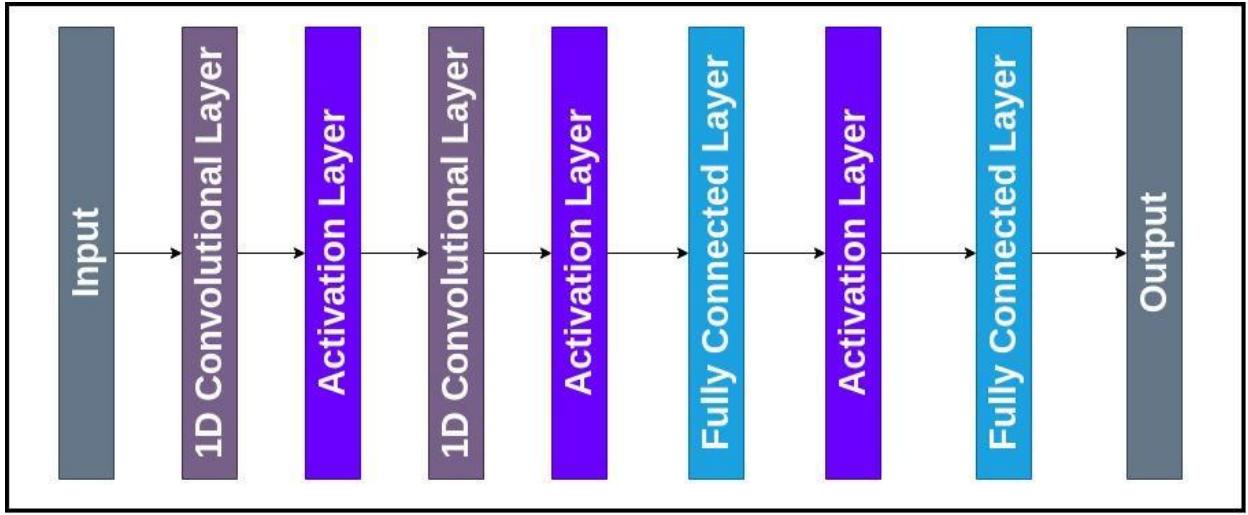

Fig. 10. The architecture of the proposed CNN-based classifier.

As is seen in Figure 10, our proposed model consists of 4 layers (two 1-dimensional convolutional layers and two fully connected layers). After each hidden layer, a non-linear activation function (ReLU) is applied to the output. To make our training process more efficient, we have experimented with several loss functions, among which Focal Loss (FL) (74) claimed better supervision of the network. In fact, it was invented to address the issue of class imbalance. FL belongs to the cost-sensitive methods which were originally introduced in the case of object detection, where the imbalance between background and salient object is often frequent. FL is a modification to the cross-entropy loss in that during the training procedure, the neural network receives more cost for wrongly predicting complex training samples.

More precisely, the cross-entropy loss function is among the most common loss functions in deep learning that originates from information theory. It is seemingly identical to the negative log-likelihood loss function, and for the binary classification problems, the binary cross-entropy loss function, denoted by $l_{BCE}$ is as follows:

$$l_{BCE}(y, \hat{y}) = -(y \log(\hat{y}) + (1-y) \log(1-\hat{y})) \quad (2)$$

Here $y, \hat{y} \in \{0,1\}^N$, where N is the number of samples, $\hat{y}$ is the predicted value, and $y$ denotes the ground truth table.

The problem with the cross-entropy loss function is that in the case of imbalance classification, the larger class overwhelms the loss by dominating the gradient (75). Hence, to obtain the Focal loss function, one can simplify and rewrite the Equation (1) in the following way:

$$CE(p, y) = \begin{cases} -\log(p), & if \ y = 1 \\ -\log(1-p), & if \ y = 0 \end{cases} \quad (3)$$

Name the probability of predicting the ground truth class $p_t$ and define $p_t$ as:

$$p_t = \begin{cases} p, & if \ y = 1 \\ 1-p, & if \ y = 0 \end{cases} \quad (4)$$

Therefore, $l_{BCE}$ can be rewritten and simplified as:

$$l_{BCE}(p, y) = CE(p_t) = -\log(p_t) \quad (5)$$

Finally, FL augments a modulating factor $\alpha(1-p_t)^\gamma$ to the binary class entropy loss function, where $\gamma > 0$ is a tunable focusing parameter which yields the following equation:

$$FL(p_t) = -\alpha_t(1 - p_t)^\gamma \log(p_t) \quad (6)$$

## 4. Experimental Results

This section comprised simulation setup, dataset description, split dataset, evaluation metrics, and classification results.

### 4.1. Simulation setup

This section includes the implementation details of our proposed methods. The tools used in this paper are listed in Table 1.

Table 1. The implementation details

| Programming Language | Python 3.9 |
|---|---|
| Deep Learning Library | PyTorch 1.9 |
| CPU | Intel® Core™ i7-10700 CPU @ 2.90GHz × 16 |
| GPU | NVIDIA Corporation GP104 [GeForce GTX 1070] |
| RAM | 64 Gb |

Moreover, in our implementation, we used the Adam algorithm to optimize the models' parameters with a learning rate of 0.001. For the loss function, FL is used with the alpha parameter set to 0.25 and gamma parameters set to 2. Further, the list of hyperparameters of the DNN and CNN models is described in detail in Tables 2 and 3.

Table 2. The list of hyperparameters of the DNN model.

| Layer # | Layer Type | Input features | Out Features |
|---|---|---|---|
| 1 | Linear (Dense) | Input shape | 64 |
| 2 | Relu | N/A | N/A |
| 3 | batch normalization | 64 | - |
| 4 | Linear (Dense) | 64 | 64 |
| 5 | Relu | N/A | N/A |
| 6 | batch normalization | 64 | - |
| 7 | dropout | rate: 0.3 | - |
| 8 | Linear (Dense) | 64 | 1 |

Table 3. The list of hyperparameters of the CNN-based model.

| Layer # | Layer Type | Input Channels | Output Channels | Kernel Size | Stride |
|---|---|---|---|---|---|
| 1 | Convolution | 1 | 16 | 3 | 1 |
| 2 | Relu | N/A | N/A | N/A | N/A |
| 3 | Convolution | 16 | 4 | 2 | 1 |
| 4 | Relu | N/A | N/A | N/A | N/A |
| 5 | Linear (Dense) | 16 | 50 | N/A | N/A |
| 6 | Relu | N/A | N/A | N/A | N/A |
| 7 | Linear (Dense) | 16 | 1 | N/A | N/A |

## 4.2. Dataset description

In order to examine our proposed methods, we used the KEEL (76) dataset repository, breast cancer, and Z-Alizadeh Sani datasets. As is depicted in Table 4, the datasets comprise different imbalanced datasets for classification tasks.

Table 4. Datasets description in detail.

| No. | Name | Attribute | # All Samples | Imbalanced Ratio |
|---|---|---|---|---|
| 1 | Wisconsin | 9 | 683 | 1.86 |
| 2 | Pima | 8 | 768 | 1.87 |
| 3 | iris0 | 4 | 150 | 2.00 |
| 4 | glass0 | 9 | 214 | 2.06 |
| 5 | glass1 | 9 | 214 | 1.82 |
| 6 | glass6 | 9 | 214 | 6.38 |
| 7 | yeast1 | 8 | 1484 | 2.46 |
| 8 | Haberman | 3 | 306 | 2.78 |
| 9 | vehicle1 | 18 | 846 | 2.90 |
| 10 | vehicle2 | 18 | 846 | 2.88 |
| 11 | vehicle3 | 18 | 846 | 2.99 |
| 12 | ecoli1 | 7 | 336 | 3.36 |
| 13 | ecoli2 | 7 | 336 | 5.46 |
| 14 | ecoli3 | 7 | 336 | 8.6 |
| 15 | new-thyroid1 | 5 | 215 | 5.14 |
| 16 | new-thyroid2 | 5 | 215 | 5.14 |
| 17 | segment0 | 19 | 2308 | 6.02 |
| 18 | yeast3 | 8 | 1484 | 8.10 |
| 19 | page-blocks0 | 10 | 5472 | 8.79 |
| 20 | yeast-2_vs_4 | 8 | 514 | 9.08 |
| 21 | penbased | 10 | 10992 | 9.41 |
| 22 | Nursery | 5 | 12690 | 2.2 |
| 23 | breast cancer | 117 | 102294 | 163.2 |
| 24 | Z-Alizadeh Sani | 55 | 303 | 2.48 |

Based on Table 4, the first column indicates the number of attributes of each dataset. The second, the sum of positive and negative samples is calculated as all samples. Also, the imbalance ratio between minority or positive and majority or negative classes is assigned in the third column. Meanwhile, the imbalance ratio is achieved by dividing negative samples into positive samples. As described in Section 3.1.3 (split dataset), the dataset was randomly shuffled and split into training and testing sets which dataset trained according to the model for 2000 epochs. The generated models were trained and evaluated for 100 runs.

### 4.4. Evaluation metrics

This section includes the elaboration of the metrics which is used to evaluate the performance of our proposed models. A fundamental classification metric tool is the Confusion Matrix. This tool is a way of demonstrating the number of correctly and incorrectly predicted samples by a classifier. It is usually a table that contains the actual and predicted state of samples compared to each other. Figure 11 depicts such a matrix for a binary classifier. This matrix includes four items, namely True Positive (TP), True Negative (TN), False Positive (FP), and False Negative (FN). The formal description of these four items is as follows:

- True Positive (TP): The number of samples that belongs to the positive class and are correctly predicted as positive by the classifier
- True Negative (TN): The number of samples that belongs to the negative class and are correctly predicted as negative by the classifier
- False Positive (FP): Number of samples that belongs to the negative class, even though they are predicted as positive by the classifier
- False Negative (FN): Number of samples that belongs to the positive class, even though they are predicted as negative by the classifier

Fig. 11. Confusion matrix.

Based on Figure 11, the classes of the minority and majority are remarked as positive and negative classes, respectively. Therefore, a confusion matrix is used to obtain performance metrics for the models on the imbalanced datasets. We utilized eight metrics such as accuracy, precision, recall, F1-score, G-Mean, specificity, AUC-ROC, and kappa for evaluating the DNN and CNN models (39, 52, 77-81).

#### 4.4.1. Accuracy

Accuracy is the ratio of the number of samples that are predicted correctly to the total of the input samples, as formulated in (7).

$$Accuracy = \frac{TP+TN}{FP+FN+TP+TN} \quad (7)$$

#### 4.4.2. Specificity

The specificity is the proportion of true-negative samples to the overall number of true-negative and false-positive samples. The specificity or True Negative Rate (TNR) of a classifier is calculated using Equation (8).

$$Specificity = \frac{TN}{TN+FP} \quad (8)$$

### 4.3.3. Recall

A recall is another measurement that shows the ratio of predicted positive samples to all the relevant samples, meaning the samples which have been actually positive. The recall is a significant metric for imbalanced datasets, demonstrating the learning accuracy of the positive class. It is calculated by Equation (9).

$$sensitivity = \frac{TP}{TP+FN} \quad (9)$$

### 4.4.4. G-Mean

The G-mean is exploited as an accuracy metric as it can gauge the accuracy rates of majority and minority classes. It is achieved by Equation (10).

$$G - Mean = \sqrt{Sensitivity * Specificity} \quad (10)$$

### 4.4.5. Precision

Precision shows how well a classifier's performance is in terms of predicting positive samples. As Equation (11) shows, it is easily calculated by dividing the number of true positives by the total number of predicted samples as positive.

$$Precision = \frac{TP}{TP+FP} \quad (11)$$

### 4.4.6. F1-Score

F1-Score, which is also called F-score or F-measure, indicates the balance which exists between recall and precision for a classifier. The closer it is to one, the more balance between precision and recall exists. F1-Score can be obtained by Equation (12).

$$F1 - Score = \frac{2TP}{2TP+FP+FN} \quad (12)$$

### 4.4.7. Kappa

The kappa metric considers the random classification model accuracy to evaluate the obtained classification accuracy. It is an important metric that indicates whether the accuracy of the classifier is at the level of reliability. The values of the Kappa are between -1 to 1. On the other hand, three reliability levels of Kappa have been exploited to assess the accuracy are as follows:
1. Kappa >= 0.75: Robust consistency, high reliable accuracy.
2. 0.4 <= Kappa <0.75: the accuracy's reliance level is generally.
3. Kappa < 0.4: Accuracy is unreliable.
The kappa formula has been specified in (13).

$$Kappa = \frac{Accuracy-randm}{1-random} \quad (13)$$

### 4.4.8. AUC-ROC

The AUC-ROC is a crucial measurement to evaluate the performance of generated classification models. A ROC plot represents the tradeoff between true positives and false positives, which actually indicates the correlation between specificity and recall. Also, AUC specifies the amount of separability power of the classifier. The AUC range is from 0 to 1. Therefore, the higher the AUC means the model has better performance at recognizing the minority and majority classes.

## 4.5. Classification results

In this section, we demonstrate our experimental results based on the evaluation metrics such as accuracy, precision, recall, F1-score, G-Mean, specificity, AUC, and kappa. The results have been elaborated by obtaining the average for each metric on three imbalanced datasets, including the KEEL repository, breast cancer, and Z-Alizadeh Sani for classification tasks.

The results are given in Tables 5-10 for six models such as SMOTE + NORM. + CNN/DNN, TL + NORM. + CNN/DNN, OSS + NORM. + CNN/DNN, NearMiss + NORM. + CNN/DNN, ROS + NORM. + CNN/DNN, and RUS + NORM. + CNN/DNN, respectively. We marked the best results in boldface.

Table 5. SMOTE + NORM. + CNN/DNN.

| Dataset | Acc (CNN/DNN) | | Pre (CNN/DNN) | | Rec (CNN/DNN) | | F1 (CNN/DNN) | | G-Mean (CNN/DNN) | | Spe (CNN/DNN) | | AUC (CNN/DNN) | | Kap (CNN/DNN) | |
|---|---|---|---|---|---|---|---|---|---|---|---|---|---|---|---|---|
| ecoli1 | 99.11 | **99.38** | 99.13 | **99.40** | 99.11 | **99.38** | 99.11 | **99.38** | 99.11 | **99.38** | 98.83 | **99.17** | 99.00 | **99.25** | 98.21 | **98.77** |
| ecoli2 | 99.60 | **99.76** | 99.60 | **99.77** | 99.60 | **99.76** | 99.60 | **99.76** | 99.60 | **99.76** | 99.44 | **99.54** | 99.03 | **99.43** | 99.19 | **99.53** |
| ecoli3 | 99.53 | **99.64** | 99.54 | **99.65** | 99.53 | **99.64** | 99.53 | **99.64** | 99.53 | **99.64** | 99.21 | **99.39** | 99.20 | **99.31** | 99.06 | **99.29** |
| ecoli-0_vs_1 | **99.91** | 99.83 | **99.92** | 99.84 | **99.91** | 99.83 | **99.91** | 99.83 | **99.91** | 99.83 | **99.93** | 99.79 | **99.27** | 99.10 | **99.83** | 99.66 |
| glass0 | **99.26** | 99.19 | **99.27** | 99.26 | **99.26** | 99.19 | **99.26** | 99.19 | **99.26** | 99.19 | **99.27** | 99.00 | 99.31 | **99.53** | 99.27 | 98.38 |
| glass1 | **99.31** | 99.23 | **99.32** | 99.25 | **99.31** | 99.23 | **99.31** | 99.23 | **99.31** | 99.23 | 99.32 | **99.64** | 99.25 | **99.34** | 99.30 | 98.46 |
| glass6 | 100.00 | 100.00 | 100.00 | 100.00 | 100.00 | 100.00 | 100.00 | 100.00 | 100.00 | 100.00 | 100.00 | 100.00 | 100.00 | 100.00 | 100.00 | 100.00 |
| Haberman | **95.60** | 95.44 | 95.61 | **96.30** | 95.60 | **95.77** | 95.60 | **96.03** | 95.60 | 96.10 | 95.61 | **96.45** | 97.03 | **98.05** | 95.00 | **96.01** |
| iris0 | 100.00 | 100.00 | 100.00 | 100.00 | 100.00 | 100.00 | 100.00 | 100.00 | 100.00 | 100.00 | 100.00 | 100.00 | 99.99 | 99.99 | 100.00 | 100.00 |
| new-thyroid1 | 99.95 | **99.96** | 99.96 | 99.96 | 99.95 | **99.96** | 99.60 | **99.96** | 99.96 | 99.96 | 99.95 | 99.92 | 99.90 | **99.95** | 99.96 | 99.92 |
| new-thyroid2 | 99.95 | **99.97** | 99.96 | **99.97** | 99.95 | **99.97** | 99.95 | **99.97** | 99.95 | **99.97** | 99.96 | 99.94 | 99.38 | **99.61** | 99.94 | 99.94 |
| page-blocks0 | **99.42** | 99.39 | **99.42** | 99.39 | **99.42** | 99.39 | **99.42** | 99.39 | **99.42** | 99.39 | **99.38** | 99.25 | 99.27 | **99.28** | 99.23 | **98.97** |
| pima | **99.35** | 99.15 | **99.36** | 99.15 | **99.35** | 99.14 | **99.35** | 99.15 | **99.35** | 99.21 | 99.30 | **99.37** | **99.81** | 99.25 | **99.46** | 99.03 |
| segment0 | 99.99 | 99.99 | 99.99 | 99.99 | 99.99 | 99.99 | 99.99 | 99.99 | 99.99 | 99.99 | 99.99 | 99.99 | **99.67** | 99.31 | 99.99 | 99.99 |
| vehicle0 | 99.95 | **99.97** | 99.95 | **99.97** | 99.95 | **99.97** | 99.95 | **99.97** | 99.95 | **99.97** | 99.89 | **99.95** | 99.83 | **99.60** | 99.88 | **99.95** |
| vehicle1 | **99.81** | 99.76 | **99.81** | 99.76 | **99.81** | 99.76 | **99.81** | 99.76 | **99.81** | 99.76 | **99.87** | 99.67 | **99.90** | 99.15 | **99.90** | 99.52 |
| vehicle2 | **99.99** | 99.99 | **99.99** | 99.99 | **99.99** | 99.99 | **99.99** | 99.99 | **99.99** | 99.99 | **99.99** | 99.99 | **99.84** | 99.57 | 99.97 | **99.98** |
| vehicle3 | **99.85** | 99.74 | **99.86** | 99.75 | **99.85** | 99.74 | **99.85** | 99.74 | **99.85** | 99.74 | **99.82** | 99.62 | **99.80** | 99.20 | **99.91** | 99.48 |
| wisconsin | 99.80 | **99.84** | 99.81 | **99.85** | 99.80 | **99.84** | 99.80 | **99.84** | 99.80 | **99.84** | **99.80** | 99.75 | 99.46 | **99.51** | **99.64** | 99.60 |

| Dataset | | | | | | | | | | | | | | | |
|---|---|---|---|---|---|---|---|---|---|---|---|---|---|---|---|
| yeast1 | **98.98** | 98.71 | **98.98** | 98.72 | **98.98** | 98.71 | **98.98** | 98.71 | **98.98** | 98.71 | **98.98** | 99.81 | **98.80** | 98.62 | **98.67** | 98.35 |
| yeast3 | **100.00** | **100.00** | **100.00** | **100.00** | **100.00** | **100.00** | **100.00** | **100.00** | **100.00** | **100.00** | **100.00** | **100.00** | **100.00** | **100.00** | 99.90 | **99.94** |
| yeast-2_vs_4 | **99.95** | 99.91 | **99.96** | 99.92 | **99.95** | 99.91 | **99.95** | 99.91 | **99.95** | 99.91 | **99.94** | 99.92 | **99.80** | 99.45 | **99.73** | 99.67 |
| penbased | **99.99** | **99.99** | **99.99** | **99.99** | **99.99** | **99.99** | **99.99** | **99.99** | **99.99** | **99.99** | **99.88** | 99.81 | **99.98** | 99.96 | **99.99** | **99.99** |
| nursery | **88.71** | 87.42 | **88.72** | 87.43 | **88.71** | 87.42 | **88.71** | 87.42 | **88.71** | 87.42 | **88.62** | 87.30 | **90.00** | 89.92 | **88.40** | 87.30 |
| breast cancer | **99.67** | 98.84 | **99.68** | 98.85 | **99.67** | 98.84 | **99.67** | 98.84 | **99.67** | 98.84 | **99.46** | 98.67 | **99.42** | 99.14 | **99.48** | 98.62 |
| Z-Alizadeh Sani | **98.57** | 97.91 | **98.58** | 97.92 | **98.57** | 97.91 | **98.57** | 97.91 | **98.57** | 97.85 | **98.42** | 97.84 | **99.14** | 99.02 | **98.21** | 98.04 |
| Average | **99.08** | 98.96 | **99.09** | 99.00 | **99.08** | 98.97 | **99.09** | 98.98 | **99.08** | 98.98 | **99.03** | 98.99 | **99.08** | 99.02 | **98.92** | 98.78 |

Table 6. TL + NORM. + CNN/DNN.

| Dataset | Acc (CNN/DNN) | | Pre (CNN/DNN) | | Rec (CNN/DNN) | | F1 (CNN/DNN) | | G-Mean (CNN/DNN) | | Spe (CNN/DNN) | | AUC (CNN/DNN) | | Kap (CNN/DNN) | |
|---|---|---|---|---|---|---|---|---|---|---|---|---|---|---|---|---|
| ecoli1 | **99.05** | 98.98 | **99.06** | 98.99 | **99.05** | 98.98 | **99.05** | 98.98 | **99.05** | 98.99 | **99.00** | 98.98 | **99.02** | 98.80 | **99.10** | 98.96 |
| ecoli2 | 99.51 | **99.76** | 99.52 | **99.77** | 99.51 | **99.76** | 99.51 | **99.76** | 99.51 | **99.76** | 99.45 | **99.54** | 99.03 | **99.70** | 99.19 | **99.53** |
| ecoli3 | 99.23 | **99.74** | 99.24 | **99.75** | 99.23 | **99.74** | 99.23 | **99.74** | 99.23 | **99.74** | 99.20 | **99.41** | 99.05 | **99.52** | 99.00 | **99.30** |
| ecoli-0_vs_1 | **99.81** | 99.75 | **99.82** | 99.76 | **99.81** | 99.75 | **99.81** | 99.75 | **99.81** | 99.75 | **99.82** | 99.63 | **99.60** | 99.17 | **99.80** | 99.61 |
| glass0 | **99.05** | 98.98 | **99.06** | 98.99 | **99.05** | 98.98 | **99.05** | 98.98 | **99.05** | 98.99 | **99.02** | 99.00 | **99.00** | 98.90 | **99.02** | 98.38 |
| glass1 | **98.83** | 98.32 | **98.84** | 98.33 | **98.83** | 98.32 | **98.83** | 98.32 | **98.83** | 98.32 | **98.83** | 98.42 | 99.15 | **99.32** | **98.80** | 98.32 |
| glass6 | **100.00** | **100.00** | **100.00** | **100.00** | **100.00** | **100.00** | **100.00** | **100.00** | **100.00** | **100.00** | **100.00** | **100.00** | **100.00** | **100.00** | **100.00** | **100.00** |
| haberman | 94.20 | **95.24** | 94.21 | **95.25** | 94.20 | **95.25** | 94.20 | **95.25** | 94.20 | **95.24** | 94.25 | **95.25** | **97.23** | **97.23** | 95.00 | **96.35** |
| iris0 | **100.00** | **100.00** | **100.00** | **100.00** | **100.00** | **100.00** | **100.00** | **100.00** | **100.00** | **100.00** | **100.00** | **100.00** | **100.00** | **100.00** | **100.00** | **100.00** |
| new-thyroid1 | 99.75 | **99.78** | 99.76 | **99.79** | 99.75 | **99.78** | 99.75 | **99.78** | 99.75 | **99.78** | 99.76 | 99.74 | **99.80** | 99.60 | 99.54 | **99.61** |

| Dataset | | | | | | | | | | | | | | | | |
|---|---|---|---|---|---|---|---|---|---|---|---|---|---|---|---|---|
| new-thyroid2 | **99.89** | 99.85 | **99.90** | 99.86 | **99.89** | 99.85 | **99.89** | 99.85 | **99.89** | 99.85 | **99.88** | 99.82 | 99.51 | **99.74** | **99.80** | 99.74 |
| page-blocks0 | 98.62 | **99.39** | 98.63 | **99.39** | 98.62 | **99.39** | 98.62 | 99.39 | 98.62 | **99.39** | 98.60 | **99.25** | **99.37** | 99.28 | **99.21** | 98.97 |
| pima | **99.27** | 99.10 | **99.28** | 99.11 | **99.27** | 99.10 | **99.27** | 99.10 | **99.27** | 99.10 | **99.24** | 99.10 | **99.12** | 99.10 | **99.05** | 99.04 |
| segment0 | **99.99** | **99.99** | **99.99** | **99.99** | **99.99** | **99.99** | **99.99** | **99.99** | **99.99** | **99.99** | **99.99** | **99.99** | **99.67** | 99.31 | **99.99** | **99.99** |
| vehicle0 | 99.90 | **99.97** | 99.91 | **99.97** | 99.90 | **99.97** | 99.90 | **99.97** | 99.90 | **99.97** | 99.90 | **99.95** | **99.85** | 99.60 | 99.87 | **99.95** |
| vehicle1 | 99.70 | **99.75** | 99.71 | **99.76** | 99.70 | **99.75** | 99.70 | **99.75** | 99.70 | **99.75** | **99.77** | 99.65 | **99.78** | 99.66 | **99.68** | 99.50 |
| vehicle2 | **99.99** | **99.99** | **99.99** | **99.99** | **99.99** | **99.99** | **99.99** | **99.99** | **99.99** | **99.99** | **99.99** | **99.99** | **99.83** | 99.63 | **99.97** | **99.98** |
| vehicle2 | **99.80** | 99.76 | **99.81** | 99.75 | **99.80** | 99.74 | **99.81** | 99.74 | **99.80** | 99.74 | **99.80** | 99.62 | **99.81** | 99.20 | **99.91** | 99.56 |
| wisconsin | 99.50 | **99.84** | 99.51 | **99.85** | 99.50 | **99.84** | 99.50 | **99.84** | 99.50 | **99.84** | 99.50 | **99.75** | **99.51** | **99.51** | 99.40 | **99.60** |
| yeast1 | **98.98** | 98.71 | **98.98** | 98.72 | **98.98** | 98.71 | **98.98** | 98.71 | **98.98** | 98.71 | 98.98 | **99.81** | **98.80** | 98.62 | **98.67** | 98.35 |
| yeast3 | 100.00 | 100.00 | 100.00 | 100.00 | 100.00 | 100.00 | 100.00 | 100.00 | 100.00 | 100.00 | 100.00 | 100.00 | 100.00 | 100.00 | 99.92 | 99.93 |
| yeast-2_vs_4 | 99.85 | **99.89** | 99.86 | **99.90** | 99.85 | **99.89** | 99.85 | **99.89** | 99.85 | **99.89** | 99.84 | **99.88** | **99.70** | 99.62 | **99.64** | 99.60 |
| penbased | **99.99** | **99.99** | **99.99** | **99.99** | **99.99** | **99.99** | **99.99** | **99.99** | **99.99** | **99.99** | **99.81** | 99.77 | **99.90** | 99.85 | **99.99** | **99.99** |
| Nursery | **89.84** | 88.21 | **89.85** | 88.22 | **89.84** | 88.21 | **89.84** | 88.21 | **89.84** | 88.21 | **89.70** | 88.11 | **91.05** | 90.95 | **89.20** | 88.00 |
| breast cancer | **99.05** | 98.70 | **99.06** | 98.71 | **99.05** | 98.70 | **99.05** | 98.70 | **99.05** | 98.70 | **98.90** | 98.55 | **99.26** | 99.07 | **98.89** | 98.57 |
| Z-Alizadeh Sani | **98.34** | 97.79 | **98.35** | 97.80 | **98.34** | 97.79 | **98.34** | 97.79 | **98.34** | 97.79 | **98.29** | 97.62 | **98.80** | 98.34 | **98.11** | 97.88 |
| Average | 98.92 | **98.95** | **98.93** | 98.90 | **98.92** | 98.90 | **98.92** | 98.90 | **98.92** | 98.90 | **98.90** | 98.87 | **99.07** | 98.98 | **98.87** | 98.79 |

Table 7. OSS + NORM. + CNN/DNN.

| Dataset | Acc (CNN/DNN) | | Pre (CNN/DNN) | | Rec (CNN/DNN) | | F1 (CNN/DNN) | | G-Mean (CNN/DNN) | | Spe (CNN/DNN) | | AUC (CNN/DNN) | | Kap (CNN/DNN) | |
|---|---|---|---|---|---|---|---|---|---|---|---|---|---|---|---|---|
| ecoli1 | **98.70** | 98.55 | **98.71** | 98.56 | **98.70** | 98.55 | **98.70** | 98.55 | **98.70** | 98.55 | **98.65** | 98.50 | 99.12 | **99.24** | **98.75** | 98.45 |
| ecoli2 | 98.41 | **98.62** | 98.42 | **98.63** | 98.41 | **98.62** | 98.41 | **98.62** | 98.41 | **98.62** | 98.36 | **98.60** | 99.13 | **99.50** | 98.25 | **98.54** |
| ecoli3 | 99.20 | **99.63** | 99.21 | **99.64** | 99.20 | **99.63** | 99.20 | **99.63** | 99.20 | **99.63** | 99.15 | **99.52** | 99.26 | **99.60** | 99.03 | **99.31** |
| ecoli-0_vs_1 | 99.59 | **99.80** | 99.60 | **99.81** | 99.59 | **99.80** | 99.59 | **99.80** | 99.59 | **99.80** | 99.48 | **99.74** | 99.40 | **99.41** | **99.58** | 99.21 |
| glass0 | **99.15** | 99.10 | **99.16** | 99.11 | **99.15** | 99.10 | **99.15** | 99.10 | **99.15** | 99.10 | **99.10** | 99.10 | 99.01 | **99.18** | 98.99 | **99.00** |
| glass1 | **98.78** | 98.40 | **98.79** | 98.41 | **98.78** | 98.40 | **98.78** | 98.40 | **98.78** | 98.40 | **98.62** | 98.38 | **99.05** | 98.99 | **98.98** | 98.84 |
| **glass6** | 100.00 | 100.00 | 100.00 | 100.00 | 100.00 | 100.00 | 100.00 | 100.00 | 100.00 | 100.00 | 100.00 | 100.00 | 100.00 | 100.00 | 100.00 | 100.00 |
| haberman | 94.50 | 95.14 | 94.51 | 95.15 | 94.50 | 95.15 | 94.50 | 95.15 | 94.50 | 95.14 | 94.46 | 95.30 | 98.00 | 97.20 | 94.89 | 96.30 |
| **iris0** | 100.00 | 100.00 | 100.00 | 100.00 | 100.00 | 100.00 | 100.00 | 100.00 | 100.00 | 100.00 | 100.00 | 100.00 | 100.00 | 100.00 | 100.00 | 100.00 |
| new-thyroid1 | **99.70** | 99.63 | **99.71** | 99.64 | **99.70** | 99.63 | **99.70** | 99.63 | **99.70** | 99.63 | **99.62** | 99.60 | **99.77** | 99.05 | **99.80** | 99.61 |
| new-thyroid2 | **99.80** | **99.80** | **99.81** | **99.81** | **99.80** | **99.80** | **99.80** | **99.80** | **99.80** | **99.80** | **99.74** | **99.74** | 99.20 | **99.55** | 99.22 | **99.74** |
| page-blocks0 | 98.43 | **99.21** | 98.44 | **99.22** | 98.43 | **99.21** | 98.43 | **99.21** | 98.43 | **99.21** | 98.21 | **99.15** | 98.50 | 99.40 | **99.20** | 98.84 |
| **pima** | **99.50** | 99.11 | **99.51** | 99.12 | **99.50** | 99.11 | **99.50** | 99.11 | **99.50** | 99.11 | **99.48** | 99.09 | **99.32** | 99.05 | **99.14** | 99.09 |
| segment0 | **99.99** | **99.99** | **99.99** | **99.99** | **99.99** | **99.99** | **99.99** | **99.99** | **99.99** | **99.99** | **99.99** | **99.99** | **99.38** | 99.20 | **99.99** | **99.99** |
| vehicle0 | 99.70 | **99.80** | 99.71 | **99.81** | 99.70 | **99.80** | 99.70 | **99.80** | 99.70 | **99.80** | 99.70 | **99.74** | **99.63** | 99.62 | **99.87** | 99.70 |
| vehicle1 | 99.30 | **99.48** | 99.31 | **99.49** | 99.30 | **99.48** | 99.30 | **99.48** | 99.30 | **99.48** | 99.26 | **99.40** | 99.12 | **99.20** | **99.40** | 99.20 |
| vehicle2 | **99.99** | **99.99** | **99.99** | **99.99** | **99.99** | **99.99** | **99.99** | **99.99** | **99.99** | **99.99** | **99.99** | **99.99** | **99.60** | 99.40 | **99.95** | **99.95** |
| vehicle2 | 99.60 | **99.70** | 99.61 | **99.71** | 99.60 | **99.70** | 99.60 | **99.70** | 99.60 | **99.70** | **99.58** | 99.56 | **99.74** | 99.22 | **99.54** | 99.47 |
| wisconsin | 99.50 | **99.75** | 99.51 | **99.76** | 99.50 | **99.75** | 99.50 | **99.75** | 99.50 | **99.75** | 99.50 | **99.70** | 99.51 | **99.60** | 99.40 | **99.51** |
| yeast1 | **98.85** | 98.70 | **98.86** | 98.71 | **98.85** | 98.71 | **98.85** | 98.71 | **98.85** | 98.71 | 98.81 | **99.69** | **98.40** | 98.31 | **98.52** | 98.22 |

| dataset | | | | | | | | | | | | | | | | |
|---|---|---|---|---|---|---|---|---|---|---|---|---|---|---|---|---|
| yeast3 | **100.00** | **100.00** | **100.00** | **100.00** | **100.00** | **100.00** | **100.00** | **100.00** | **100.00** | **100.00** | **100.00** | **100.00** | **100.00** | **100.00** | **99.81** | 99.70 |
| yeast-2_vs_4 | 99.85 | **99.89** | 99.86 | **99.90** | 99.85 | **99.89** | 99.85 | **99.89** | 99.85 | **99.89** | 99.84 | **99.88** | **99.70** | 99.62 | **99.64** | 99.60 |
| penbased | **99.99** | **99.99** | **99.99** | **99.99** | **99.99** | **99.99** | **99.99** | **99.99** | **99.99** | **99.99** | 99.76 | 99.69 | 99.84 | 99.73 | **99.99** | **99.99** |
| nursery | **89.27** | 88.88 | **89.28** | 88.89 | **89.27** | 88.88 | **89.27** | 88.88 | **89.27** | 88.88 | 89.15 | **88.65** | 90.31 | **91.23** | 89.06 | **88.70** |
| breast cancer | **99.15** | 98.81 | **99.16** | 98.82 | **99.15** | 98.81 | **99.15** | 98.81 | **99.15** | 98.81 | **99.01** | 98.69 | **99.19** | 98.99 | **99.03** | 98.73 |
| Z-Alizadeh Sani | **97.56** | 97.43 | **97.57** | 97.44 | **97.56** | 97.43 | **97.56** | 97.43 | **97.56** | 97.43 | **97.27** | 97.21 | **98.24** | 98.20 | **97.20** | 97.10 |
| Average | 98.78 | **98.82** | 98.79 | **98.83** | 98.78 | **98.82** | 98.78 | **98.82** | 98.78 | **98.82** | 98.72 | **98.80** | 98.93 | **98.94** | **98.73** | 98.72 |

Table 8. NearMiss + NORM. + CNN/DNN.

| Dataset | Acc (CNN/DNN) | | Pre (CNN/DNN) | | Rec (CNN/DNN) | | F1 (CNN/DNN) | | G-Mean (CNN/DNN) | | Spe (CNN/DNN) | | AUC (CNN/DNN) | | Kap (CNN/DNN) | |
|---|---|---|---|---|---|---|---|---|---|---|---|---|---|---|---|---|
| ecoli1 | 99.16 | **99.20** | 99.17 | **99.21** | 99.16 | **99.20** | 99.16 | **99.20** | 99.16 | **99.20** | 99.10 | **99.15** | **99.41** | 99.31 | **99.11** | 99.11 |
| ecoli2 | 99.60 | **99.88** | 99.61 | **99.89** | 99.60 | **99.88** | 99.60 | **99.88** | 99.60 | **99.88** | 99.52 | **99.80** | 99.23 | **99.55** | 99.33 | **99.88** |
| ecoli3 | 99.33 | **99.74** | 99.34 | **99.75** | 99.33 | **99.74** | 99.33 | **99.74** | 99.33 | **99.74** | 99.29 | **99.41** | 99.15 | **99.52** | 99.02 | **99.30** |
| ecoli-0_vs_1 | 99.60 | **99.78** | 99.61 | **99.79** | 99.60 | **99.78** | 99.60 | **99.78** | 99.60 | **99.78** | 99.54 | **99.74** | 99.35 | **99.37** | **99.57** | 99.54 |
| glass0 | 99.20 | **99.25** | 99.21 | **99.26** | 99.20 | **99.25** | 99.20 | **99.25** | 99.20 | **99.25** | 99.13 | **99.18** | 99.11 | **99.18** | **99.18** | 99.06 |
| glass1 | 99.15 | **99.20** | 99.16 | **99.21** | 99.15 | **99.20** | 99.15 | **99.20** | 99.15 | **99.20** | 99.10 | **99.17** | 99.18 | **99.25** | 99.06 | **99.14** |
| glass6 | **100.00** | 100.00 | 100.00 | 100.00 | 100.00 | 100.00 | 100.00 | 100.00 | 100.00 | 100.00 | 100.00 | 100.00 | 100.00 | 100.00 | 100.00 | 100.00 |
| haberman | 94.80 | **95.36** | 94.81 | **95.37** | 94.80 | **95.37** | 94.80 | **95.37** | 94.80 | **95.37** | 94.62 | **95.32** | **98.00** | 97.25 | 95.00 | **95.35** |
| iris0 | 100.00 | 100.00 | 100.00 | 100.00 | 100.00 | 100.00 | 100.00 | 100.00 | 100.00 | 100.00 | 100.00 | 100.00 | 100.00 | 100.00 | 100.00 | 100.00 |
| new-thyroid1 | 99.63 | **99.65** | 99.64 | **99.66** | 99.63 | **99.65** | 99.63 | **99.65** | 99.63 | **99.64** | 99.59 | **99.61** | 99.74 | **99.80** | 99.50 | **99.55** |
| new-thyroid2 | **99.90** | 99.87 | **99.91** | 99.88 | **99.90** | 99.87 | **99.90** | 99.87 | **99.90** | 99.87 | 99.84 | **99.87** | 99.50 | **99.70** | **99.87** | 99.78 |
| page-blocks0 | 99.20 | **99.43** | 99.21 | **99.44** | 99.20 | **99.43** | 99.20 | **99.43** | 99.20 | **99.43** | 99.15 | **99.39** | **99.35** | 99.24 | 99.11 | **99.24** |
| pima | **99.30** | 99.15 | **99.31** | 99.16 | **99.30** | 99.15 | **99.30** | 99.15 | **99.30** | 99.15 | **99.25** | 99.11 | **99.25** | 99.11 | **99.15** | 99.08 |
| segment0 | **99.99** | 99.99 | **99.99** | 99.99 | **99.99** | 99.99 | **99.99** | 99.99 | **99.99** | 99.99 | **99.99** | 99.99 | **99.70** | 99.51 | **99.99** | 99.99 |
| vehicle0 | **99.99** | 99.99 | **99.99** | 99.99 | **99.99** | 99.99 | **99.99** | 99.99 | **99.99** | 99.99 | **99.99** | 99.99 | **99.78** | 99.62 | **99.99** | 99.99 |
| vehicle1 | **99.99** | 99.99 | **99.99** | 99.99 | **99.99** | 99.99 | **99.99** | 99.99 | **99.99** | 99.99 | **99.99** | 99.99 | **99.84** | 99.80 | **99.99** | 99.99 |
| vehicle2 | **99.99** | 99.99 | **99.99** | 99.99 | **99.99** | 99.99 | **99.99** | 99.99 | **99.99** | 99.99 | **99.99** | 99.99 | **99.77** | 99.64 | **99.99** | 99.99 |
| vehicle3 | **99.99** | 99.99 | **99.99** | 99.99 | **99.99** | 99.99 | **99.99** | 99.99 | **99.99** | 99.99 | **99.99** | 99.99 | 99.40 | **99.50** | **99.99** | 99.99 |

| dataset | | | | | | | | | | | | | | | | |
|---|---|---|---|---|---|---|---|---|---|---|---|---|---|---|---|---|
| wisconsin | 99.42 | **99.50** | 99.43 | **99.51** | 99.42 | **99.50** | 99.42 | **99.50** | 99.42 | **99.50** | 99.35 | **99.42** | 99.20 | **99.55** | 99.32 | **99.41** |
| yeast1 | **99.30** | 99.25 | **99.31** | 99.26 | **99.30** | 99.25 | **99.30** | 99.25 | **99.30** | 99.25 | **99.24** | 99.15 | **99.58** | 98.37 | **99.24** | 98.35 |
| yeast3 | **100.00** | 100.00 | 100.00 | 100.00 | 100.00 | 100.00 | 100.00 | 100.00 | 100.00 | 100.00 | 100.00 | 100.00 | 100.00 | 100.00 | 99.81 | 99.70 |
| yeast-2_vs_4 | 99.90 | **99.91** | 99.91 | **99.92** | 99.90 | **99.91** | 99.90 | **99.91** | 99.90 | **99.91** | 99.87 | **99.92** | 99.74 | **99.85** | 99.62 | **99.67** |
| penbased | **99.99** | 99.99 | 99.99 | 99.99 | **99.99** | 99.99 | **99.99** | 99.99 | **99.99** | 99.99 | **99.81** | 99.76 | **99.92** | 99.88 | **99.99** | 99.99 |
| nursery | **88.37** | 88.18 | **88.38** | 88.19 | **88.37** | 88.18 | **88.37** | 88.18 | **88.37** | 88.18 | **88.21** | 88.02 | **90.37** | 89.86 | **88.02** | 88.00 |
| breast cancer | **99.23** | 98.92 | **99.24** | 98.93 | **99.23** | 98.92 | **99.23** | 98.92 | **99.23** | 98.92 | **99.11** | 98.75 | **99.05** | 99.02 | **99.09** | 99.02 |
| Z-Alizadeh Sani | **98.50** | 98.24 | **98.51** | 98.25 | **98.50** | 98.24 | **98.50** | 98.24 | **98.50** | 98.10 | **98.39** | 98.05 | **98.27** | 98.10 | **98.21** | 98.13 |
| Average | 98.98 | **99.01** | 98.98 | **99.02** | 98.98 | **99.01** | 98.98 | **99.01** | 98.98 | **99.01** | 98.92 | **98.95** | **99.07** | 98.99 | 98.89 | **98.89** |

Table 9. ROS + NORM. + CNN/DNN.

| Dataset | Acc (CNN/DNN) | | Pre (CNN/DNN) | | Rec (CNN/DNN) | | F1 (CNN/DNN) | | G-Mean (CNN/DNN) | | Spe (CNN/DNN) | | AUC (CNN/DNN) | | Kap (CNN/DNN) | |
|---|---|---|---|---|---|---|---|---|---|---|---|---|---|---|---|---|
| ecoli1 | **98.52** | 98.49 | **98.53** | 98.50 | **98.52** | 98.49 | **98.52** | 98.49 | **98.52** | 98.49 | **98.34** | 98.28 | **99.08** | 98.57 | **98.87** | 98.64 |
| ecoli2 | **98.81** | 98.70 | **98.82** | 98.71 | **98.81** | 98.70 | **98.81** | 98.70 | **98.81** | 98.70 | **98.72** | 98.61 | **99.38** | 99.33 | **98.76** | 98.50 |
| ecoli3 | **98.90** | 98.64 | **98.91** | 98.65 | **98.90** | 98.64 | **98.90** | 98.64 | **98.90** | 98.64 | **98.78** | 98.56 | 99.16 | **99.22** | 99.12 | **99.21** |
| ecoli-0_vs_1 | **98.81** | 98.64 | **98.82** | 98.65 | **98.81** | 98.64 | **98.81** | 98.64 | **98.81** | 98.64 | **98.70** | 98.53 | 99.06 | **99.27** | **98.54** | 98.40 |
| glass0 | **98.30** | 98.17 | **98.31** | 98.18 | **98.30** | 98.17 | **98.30** | 98.17 | **98.30** | 98.17 | **98.19** | 98.08 | **99.25** | 99.15 | 98.11 | **99.16** |
| glass1 | **98.27** | 98.16 | **98.28** | 98.17 | **98.27** | 98.16 | **98.27** | 98.16 | **98.27** | 98.16 | **98.20** | 98.10 | 99.08 | **99.19** | **98.15** | 98.02 |
| glass6 | **100.00** | **100.00** | **100.00** | **100.00** | **100.00** | **100.00** | **100.00** | **100.00** | **100.00** | **100.00** | **100.00** | **100.00** | **100.00** | **100.00** | **100.00** | **100.00** |
| Haberman | 95.45 | **95.50** | 95.46 | **95.51** | 95.45 | **95.50** | 95.45 | **95.50** | 95.45 | **95.50** | 95.30 | **95.40** | 98.00 | **98.26** | 95.24 | **95.40** |
| iris0 | **100.00** | **100.00** | **100.00** | **100.00** | **100.00** | **100.00** | **100.00** | **100.00** | **100.00** | **100.00** | **100.00** | **100.00** | **100.00** | **100.00** | **100.00** | **100.00** |
| new-thyroid1 | 98.59 | **98.63** | 98.60 | **98.64** | 98.59 | **98.63** | 98.59 | **98.63** | 98.59 | **98.63** | 98.46 | **98.60** | 99.01 | **99.14** | 98.68 | **99.85** |
| new-thyroid2 | **98.87** | 98.79 | **98.88** | 98.80 | **98.87** | 98.79 | **98.87** | 98.79 | **98.87** | 98.79 | **98.54** | 98.50 | **99.08** | 99.01 | **98.80** | 98.72 |
| page-blocks0 | **98.40** | 98.16 | **98.41** | 98.17 | **98.40** | 98.16 | **98.40** | 98.16 | **98.40** | 98.16 | **98.26** | 98.09 | **99.12** | 99.06 | **98.60** | 98.47 |
| pima | 98.32 | **99.02** | 98.33 | **99.03** | 98.32 | **99.02** | 98.32 | **99.02** | 98.32 | **99.02** | 98.24 | **98.95** | **99.20** | 99.03 | **99.23** | 99.01 |
| segment0 | **99.99** | **99.99** | **99.99** | **99.99** | **99.99** | **99.99** | **99.99** | **99.99** | **99.99** | **99.99** | **99.99** | **99.99** | **99.35** | 99.29 | **99.99** | **99.99** |
| vehicle0 | **98.98** | 98.90 | **98.99** | 98.91 | **98.98** | 98.90 | **98.98** | 98.90 | **98.98** | 98.90 | **98.86** | 98.80 | **98.60** | 98.57 | **98.86** | 98.80 |
| vehicle1 | **98.58** | 98.44 | **98.59** | 98.45 | **98.58** | 98.44 | **98.58** | 98.44 | **98.58** | 98.44 | **98.50** | 98.38 | **99.16** | 99.10 | **98.42** | 98.37 |
| vehicle2 | **99.99** | **99.99** | **99.99** | **99.99** | **99.99** | **99.99** | **99.99** | **99.99** | **99.99** | **99.99** | **99.99** | **99.99** | **99.55** | 99.34 | **99.95** | **99.95** |
| vehicle2 | **99.31** | 99.15 | **99.32** | 99.16 | **99.31** | 99.15 | **99.31** | 99.15 | **99.31** | 99.15 | **99.23** | 99.10 | **99.21** | 99.11 | **99.20** | 99.05 |
| wisconsin | **99.72** | 99.70 | **99.73** | 99.71 | **99.72** | 99.70 | **99.72** | 99.70 | **99.72** | 99.70 | **99.60** | 99.55 | **99.23** | 99.19 | **99.50** | 99.45 |
| yeast1 | **98.94** | 98.60 | **98.95** | 98.61 | **98.94** | 98.60 | **98.94** | 98.60 | **98.94** | 98.60 | **98.80** | 98.73 | **99.03** | 99.00 | **98.90** | 98.72 |
| yeast3 | **100.00** | **100.00** | **100.00** | **100.00** | **100.00** | **100.00** | **100.00** | **100.00** | **100.00** | **100.00** | **100.00** | **100.00** | **100.00** | **100.00** | 99.91 | **99.96** |
| yeast-2_vs_4 | 98.93 | **99.00** | 98.94 | **99.00** | 98.93 | **99.00** | 98.93 | **99.00** | 98.93 | **99.00** | 98.81 | **98.90** | 98.86 | **99.10** | **98.99** | 98.92 |
| penbased | **99.99** | **99.99** | **99.99** | **99.99** | **99.99** | **99.99** | **99.99** | **99.99** | **99.99** | **99.99** | **99.73** | 99.60 | **99.89** | 99.87 | **99.99** | **99.99** |
| nursery | **88.19** | 88.03 | **88.20** | 88.04 | **88.19** | 88.03 | **88.19** | 88.03 | **88.19** | 88.03 | **88.02** | 87.93 | **90.19** | 89.95 | **87.94** | 87.83 |
| breast cancer | **99.00** | 98.72 | **99.01** | 98.73 | **99.00** | 98.72 | **99.00** | 98.72 | **99.00** | 98.72 | **98.81** | 98.61 | **98.73** | 98.52 | **98.67** | 98.43 |
| Z-Alizadeh Sani | **96.87** | 96.60 | **96.88** | 96.61 | **96.87** | 96.60 | **96.87** | 96.60 | **96.87** | 96.60 | **96.72** | 96.50 | **97.19** | 97.00 | **96.72** | 96.49 |
| Average | **98.45** | 98.38 | **98.45** | 98.39 | **98.45** | 98.38 | **98.45** | 98.38 | **98.45** | 98.38 | **98.33** | 98.29 | **98.78** | 98.74 | 98.42 | **98.43** |

Table 10. RUS + NORM. + CNN/DNN.

| Dataset | Acc (CNN/DNN) | | Pre (CNN/DNN) | | Rec (CNN/DNN) | | F1 (CNN/DNN) | | G-Mean (CNN/DNN) | | Spe (CNN/DNN) | | AUC (CNN/DNN) | | Kap (CNN/DNN) | |
|---|---|---|---|---|---|---|---|---|---|---|---|---|---|---|---|---|
| ecoli1 | **98.26** | 98.17 | **98.27** | 98.18 | **98.26** | 98.17 | **98.26** | 98.17 | **98.26** | 98.17 | **98.19** | 98.07 | **99.02** | 99.00 | **98.64** | 98.60 |
| ecoli2 | **98.51** | 98.43 | **98.52** | 98.44 | **98.51** | 98.43 | **98.51** | 98.43 | **98.51** | 98.43 | **98.39** | 98.37 | **99.13** | 99.06 | **98.71** | 98.60 |
| ecoli3 | **98.84** | 98.70 | **98.85** | 98.71 | **98.84** | 98.70 | **98.84** | 98.70 | **98.84** | 98.70 | **98.76** | 98.63 | **99.19** | 99.16 | **98.80** | 98.62 |
| ecoli-0_vs_1 | **98.77** | 98.60 | **98.78** | 98.61 | **98.77** | 98.60 | **98.77** | 98.60 | **98.77** | 98.60 | **98.70** | 98.51 | **99.03** | 99.01 | **98.65** | 98.57 |
| glass0 | **98.36** | 98.13 | **98.37** | 98.14 | **98.36** | 98.13 | **98.36** | 98.13 | **98.36** | 98.13 | **98.24** | 98.03 | **99.28** | 99.06 | **98.17** | 98.11 |
| glass1 | 98.49 | **98.53** | 98.50 | **98.54** | 98.49 | **98.53** | 98.49 | **98.53** | 98.49 | **98.53** | 98.33 | **98.44** | **99.16** | 98.39 | 98.27 | **98.45** |
| glass6 | 100.00 | 100.00 | 100.00 | 100.00 | 100.00 | 100.00 | 100.00 | 100.00 | 100.00 | 100.00 | 100.00 | 100.00 | 100.00 | 100.00 | 100.00 | 100.00 |
| haberman | **94.73** | 94.61 | **94.74** | 94.62 | **94.73** | 94.61 | **94.73** | 94.61 | **94.73** | 94.61 | **94.65** | 94.52 | **99.01** | 98.89 | **94.29** | 94.20 |
| iris0 | **99.99** | 99.99 | **99.99** | 99.99 | **99.99** | 99.99 | **99.99** | 99.99 | **99.99** | 99.99 | **99.99** | 99.99 | **99.84** | 99.70 | **99.96** | 99.95 |
| new-thyroid1 | 98.29 | **98.43** | 98.30 | **98.44** | 98.30 | **98.43** | 98.30 | **98.43** | 98.30 | **98.43** | 98.05 | **98.27** | 99.32 | **99.43** | **99.27** | 98.43 |
| new-thyroid2 | **98.80** | 98.30 | **98.81** | 98.31 | **98.80** | 98.30 | **98.80** | 98.30 | **98.80** | 98.30 | **98.69** | 98.15 | **98.99** | 98.52 | **99.01** | 98.78 |
| page-blocks0 | **98.31** | 98.14 | **98.32** | 98.15 | **98.31** | 98.14 | **98.31** | 98.14 | **98.31** | 98.14 | **98.15** | 98.04 | **98.87** | 98.45 | **98.32** | 98.14 |
| pima | **98.19** | 98.05 | **98.20** | 98.06 | **98.19** | 98.05 | **98.19** | 98.05 | **98.19** | 98.05 | **98.06** | 97.94 | **98.85** | 98.56 | **98.00** | 97.94 |
| segment0 | **99.99** | 99.99 | **99.99** | 99.99 | **99.99** | 99.99 | **99.99** | 99.99 | **99.99** | 99.99 | **99.99** | 99.99 | **99.29** | 99.21 | **99.99** | 99.99 |
| vehicle0 | **98.84** | 98.73 | **98.85** | 98.74 | **98.84** | 98.73 | **98.84** | 98.73 | **98.84** | 98.73 | 98.63 | **98.73** | **99.15** | 98.99 | **98.70** | 98.52 |
| vehicle1 | **98.46** | 98.32 | **98.47** | 98.33 | **98.46** | 98.32 | **98.46** | 98.32 | **98.46** | 98.32 | **98.32** | 98.24 | **98.21** | 98.10 | **98.24** | 98.10 |
| vehicle2 | **99.99** | 99.99 | **99.99** | 99.99 | **99.99** | 99.99 | **99.99** | 99.99 | **99.99** | 99.99 | **99.99** | 99.99 | **99.33** | 99.05 | **99.82** | 99.70 |
| vehicle3 | **99.28** | 99.13 | **99.29** | 99.14 | **99.28** | 99.13 | **99.28** | 99.13 | **99.28** | 99.13 | **99.18** | 99.05 | **99.08** | 98.99 | **99.11** | 98.94 |
| wisconsin | **99.64** | 99.48 | **99.65** | 99.49 | **99.64** | 99.48 | **99.64** | 99.48 | **99.64** | 99.48 | **99.61** | 99.34 | **99.64** | 99.27 | **99.29** | 99.16 |

| Dataset | | | | | | | | | | | | | | | | |
|---|---|---|---|---|---|---|---|---|---|---|---|---|---|---|---|---|
| yeast1 | **98.99** | 98.70 | **99.00** | 98.71 | **98.99** | 98.70 | **98.99** | 98.70 | **98.99** | 98.70 | **98.79** | 98.61 | **99.02** | 98.89 | **98.75** | 98.52 |
| yeast3 | **100.00** | 100.00 | **100.00** | 100.00 | **100.00** | 100.00 | **100.00** | 100.00 | **100.00** | 100.00 | **100.00** | 100.00 | **100.00** | 100.00 | **99.84** | 99.62 |
| yeast-2_vs_4 | **98.82** | 98.70 | **98.83** | 98.71 | **98.82** | 98.70 | **98.82** | 98.70 | **98.82** | 98.62 | **98.70** | 98.70 | **98.82** | 98.50 | **98.70** | 98.63 |
| penbased | **99.99** | 99.99 | **99.99** | 99.99 | **99.99** | 99.99 | **99.99** | 99.99 | **99.99** | 99.99 | **99.92** | 99.87 | **99.90** | 99.87 | **99.99** | 99.99 |
| nursery | **88.06** | 88.00 | **88.07** | 88.01 | **88.06** | 88.00 | **88.06** | 88.00 | **88.06** | 88.00 | **87.96** | 87.80 | **89.96** | 89.85 | **87.84** | 87.95 |
| breast cancer | **98.91** | 98.70 | **98.92** | 98.71 | **98.91** | 98.70 | **98.91** | 98.70 | **98.91** | 98.70 | **98.70** | 98.60 | **98.50** | 98.43 | **98.36** | 98.27 |
| Z-Alizadeh Sani | **96.30** | 96.15 | **96.31** | 96.16 | **96.30** | 96.15 | **96.30** | 96.15 | **96.30** | 96.15 | **96.16** | 96.02 | **96.98** | 97.00 | **96.12** | 96.01 |
| Average | **98.33** | 98.22 | **98.34** | 98.23 | **98.33** | 98.22 | **98.33** | 98.22 | **98.33** | 98.22 | **98.23** | 98.15 | **98.75** | 98.59 | **98.26** | 98.14 |

*NORM.: Normalization, Accuracy: Acc, Precision: Pre, Recall: Rec F1-Score: F1, Specificity: Spe, Kappa: Kap

According to the results obtained, the proposed SMOTE + NORM. + CNN model outperforms other models in terms of eight metrics on the datasets. For a better comparison of the results, the best average performance score is also shown in Figures 12-17. According to these Figures, the mixed SMOTE-NORM.-CNN model is superior to the other models in terms of the evaluation criteria for the imbalanced datasets. As a result, this model demonstrates the impact of using SMOTE in our CNN model so that the overall performance has been enhanced.

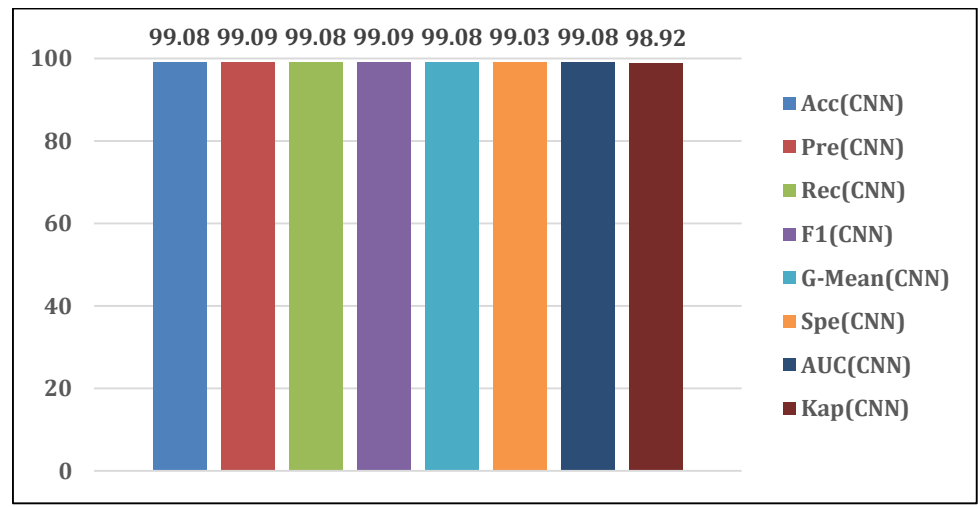

Fig. 12. The average rate of the metrics for the SMOTE + NORM. + CNN/DNN.

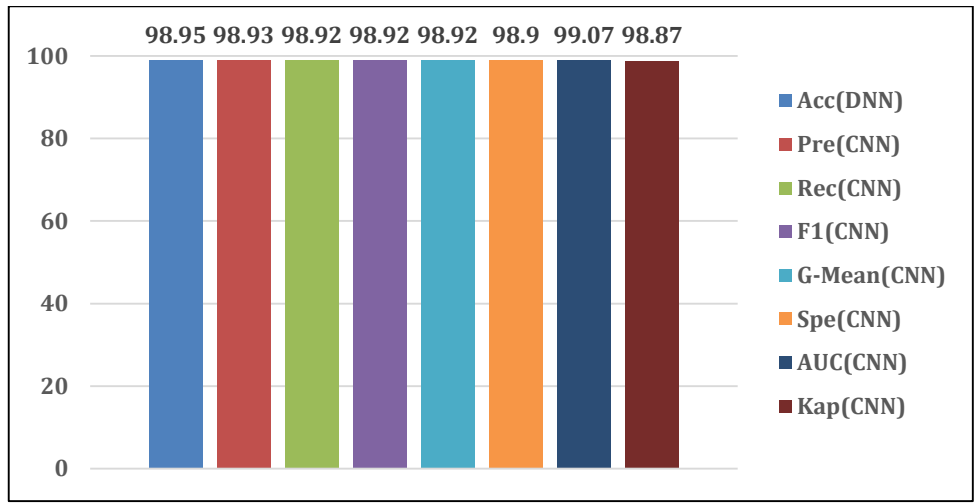

Fig. 13. The average rate of the metrics for the TL + NORM. + CNN/DNN.

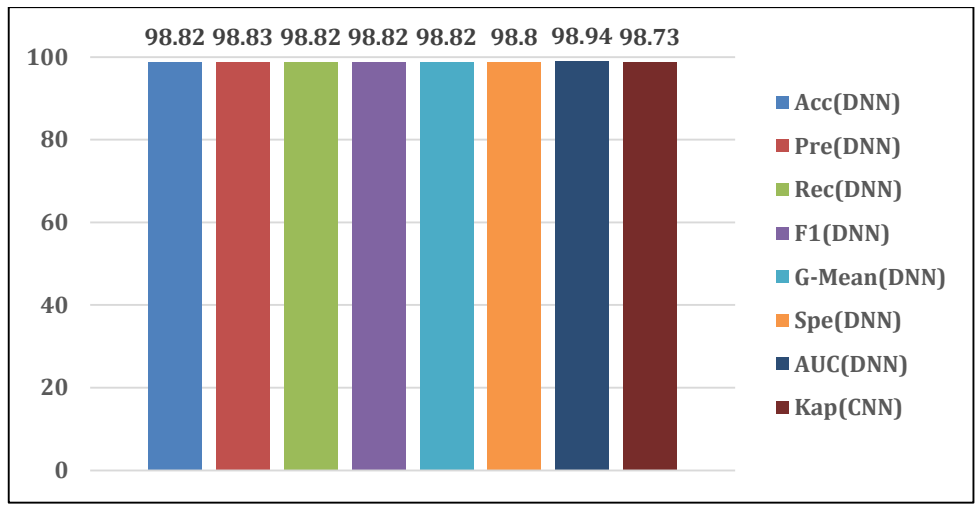

Fig. 14. The average rate of the metrics for the OSS + NORM. + CNN/DNN.

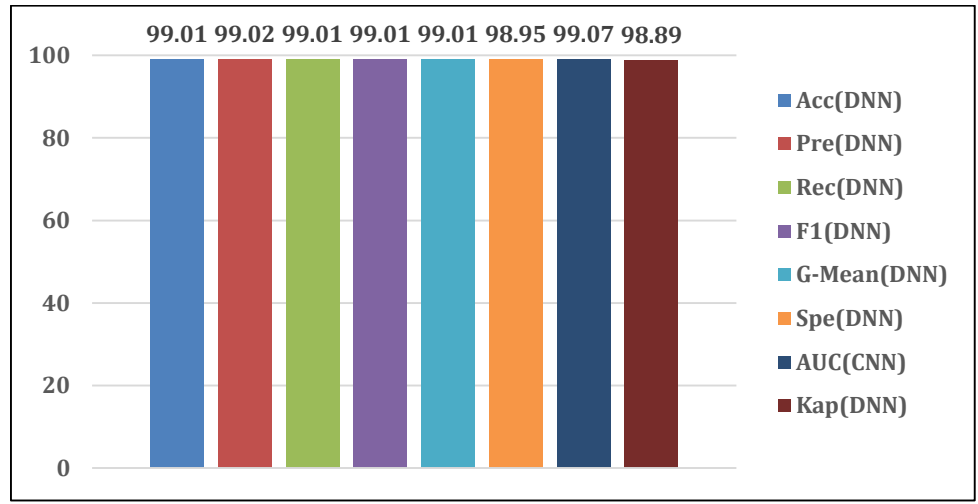

Fig. 15. The average rate of the metrics for the NearMiss + NORM. + CNN/DNN.

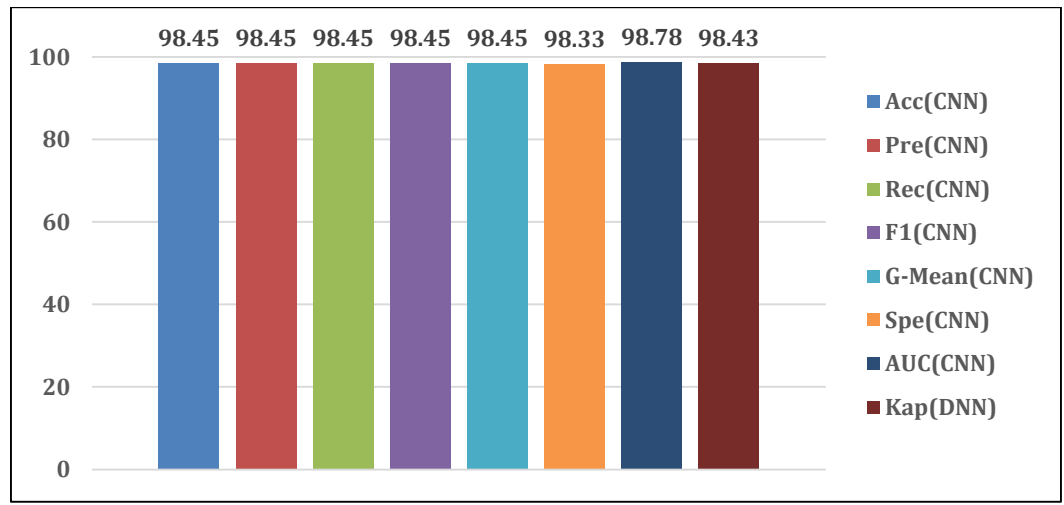

Fig. 16. The average rate of the metrics for the ROS + NORM. + CNN/DNN.

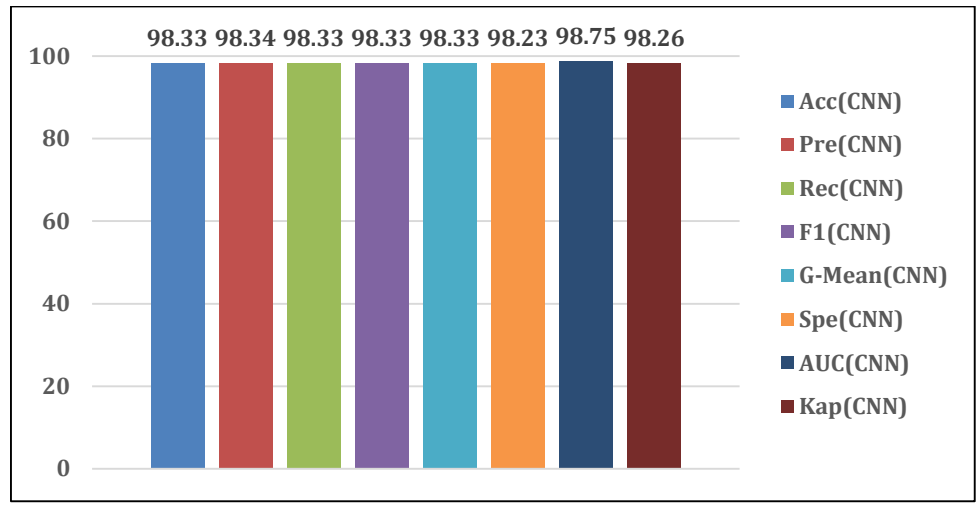

Fig. 17. The average rate of the metrics for the RUS + NORM. + CNN/DNN.

Moreover, in our experiment, the ROC plots based on the best AUC scores gained through the models are shown in Figures 18 (a)-(z) on the datasets. Due to Tables 6-11 and ROC plots, the mixed SMOTE-NORM.-CNN model has the best AUC value.

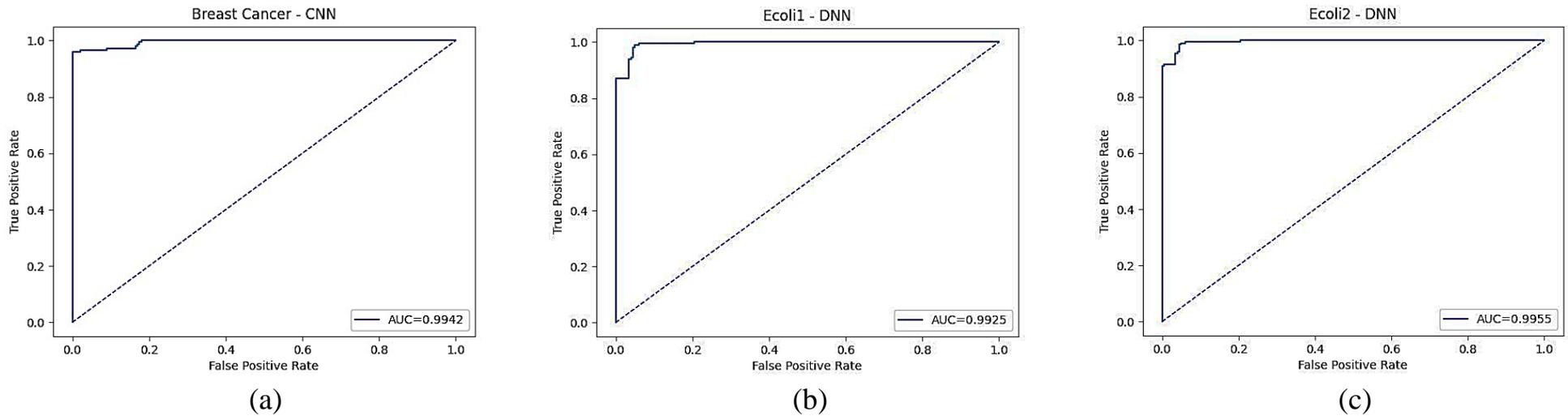

(a)          (b)          (c)

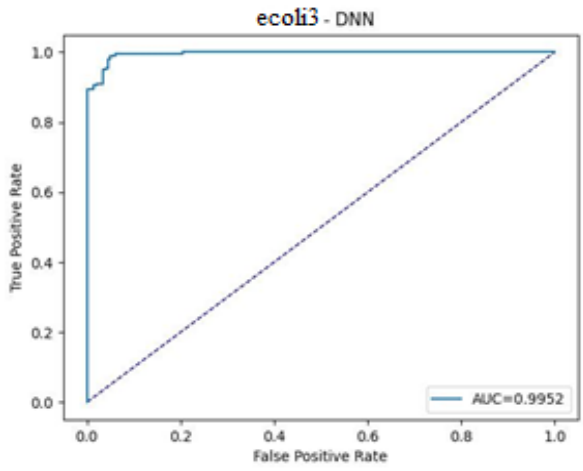
(d)
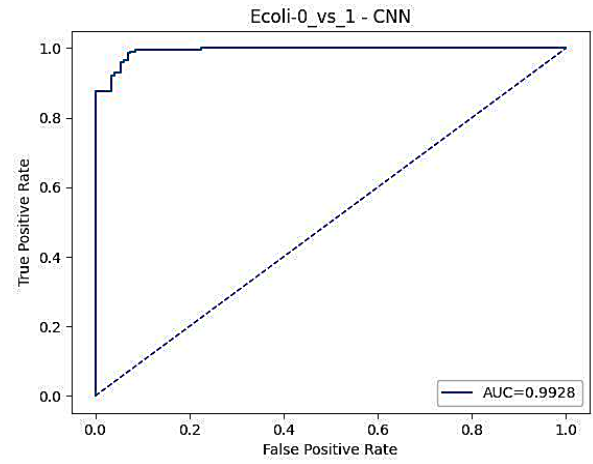
(e)
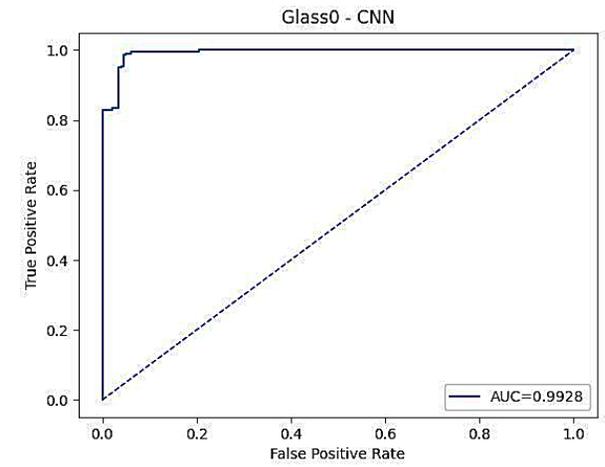
(f)
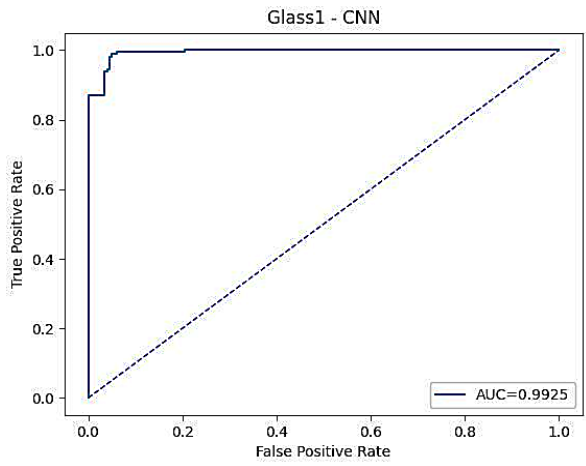
(g)
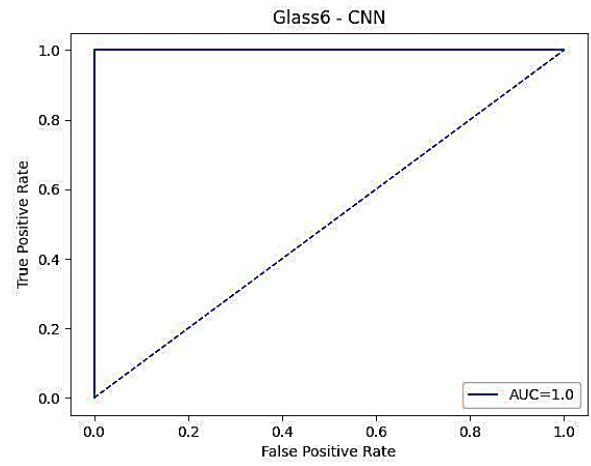
(h)
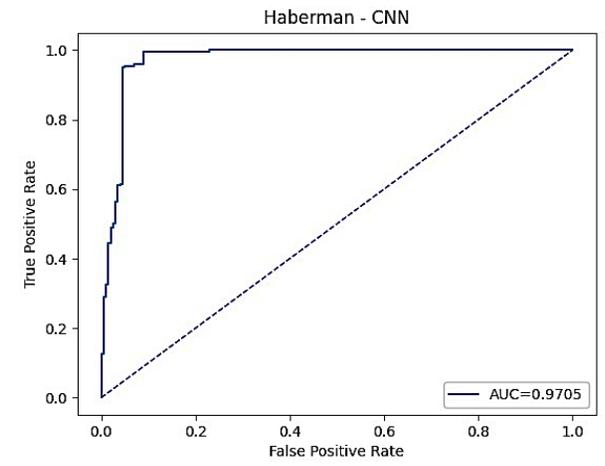
(i)

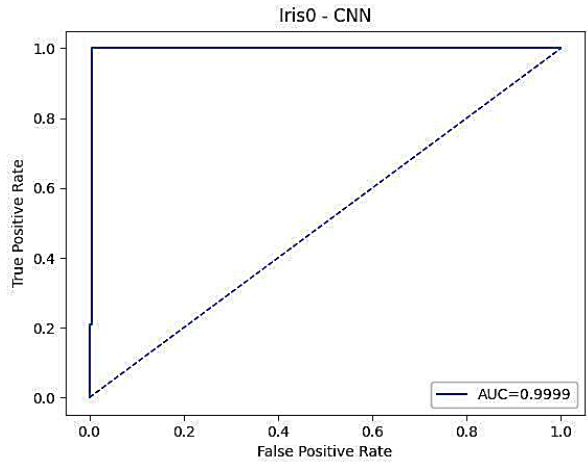
(j)
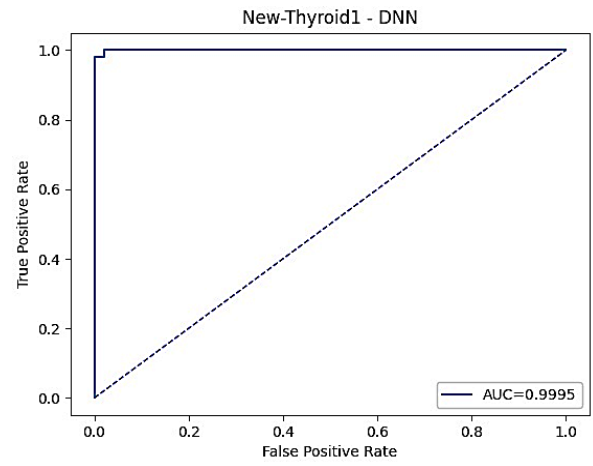
(k)
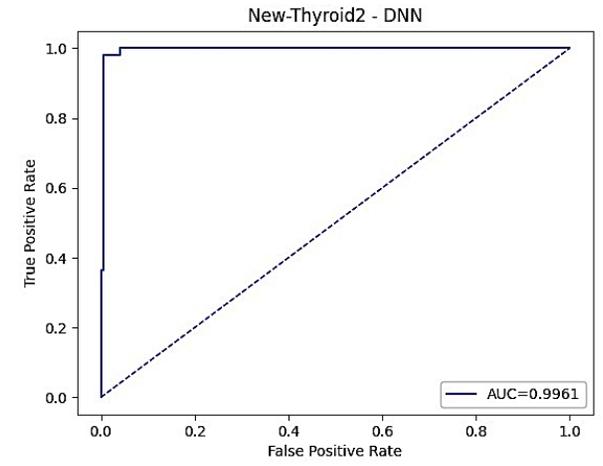
(l)
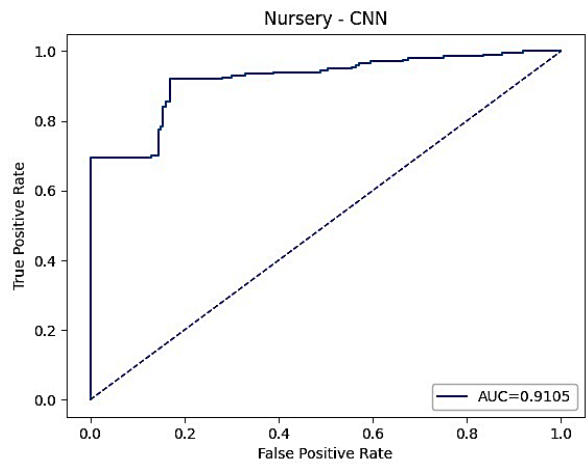
(m)
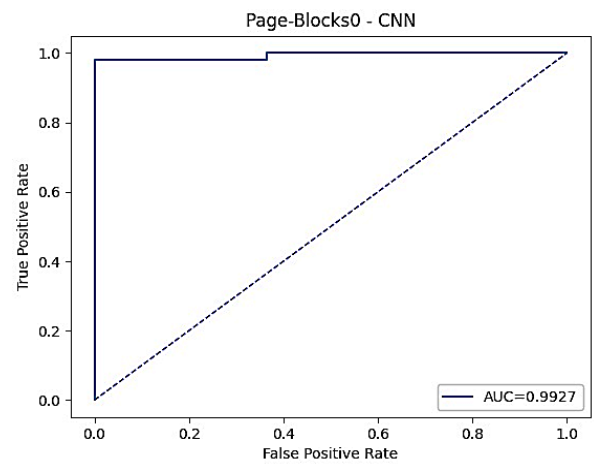
(n)
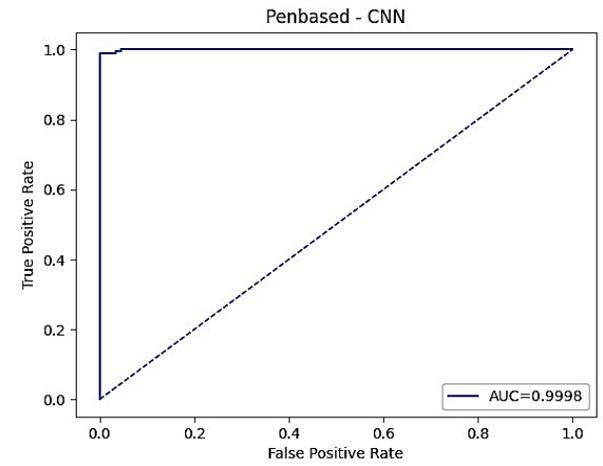
(o)

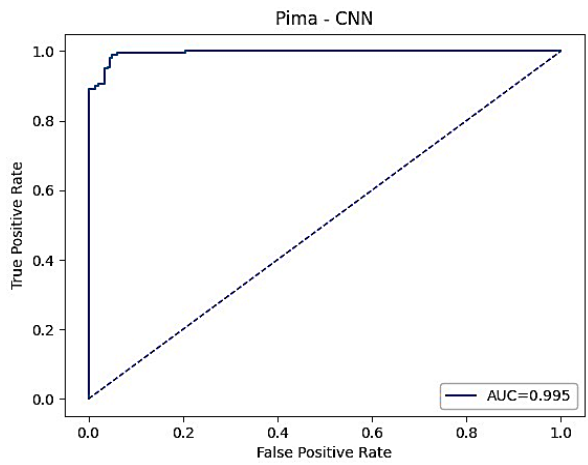

(p)

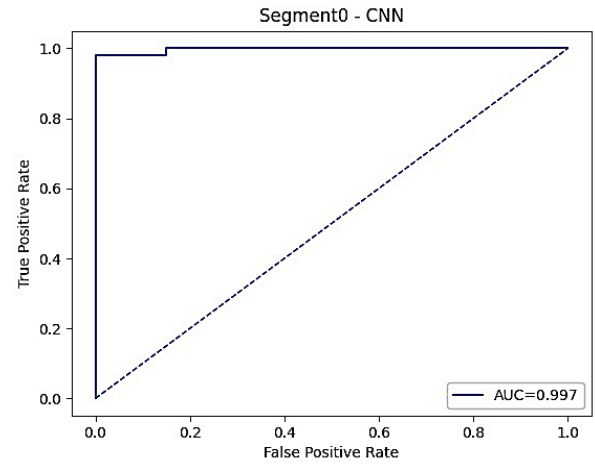

(q)

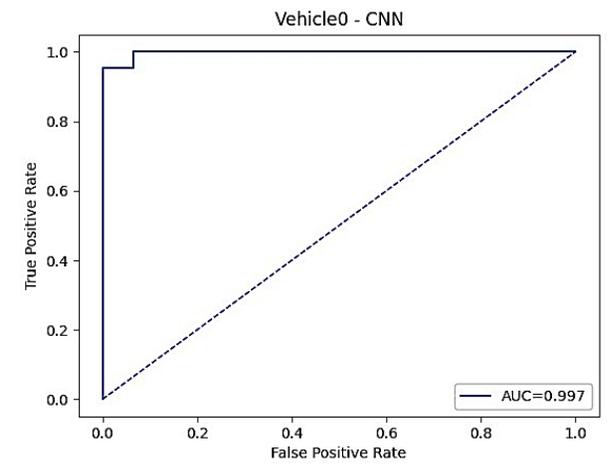

(r)

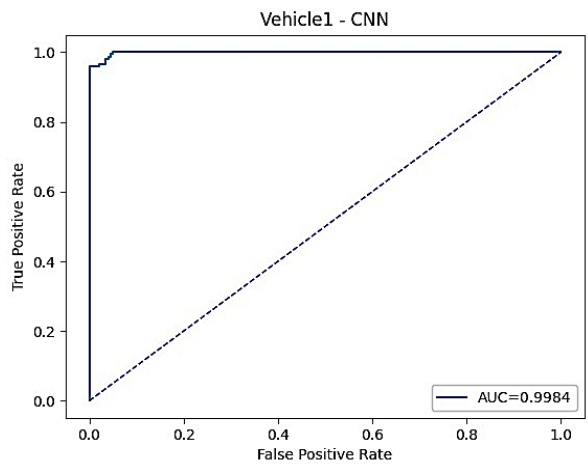

(s)

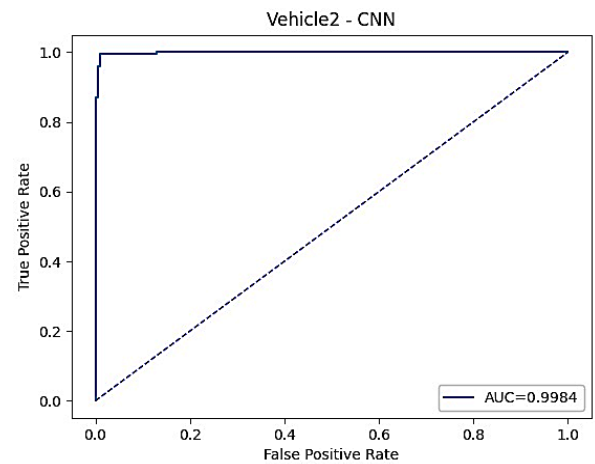

(t)

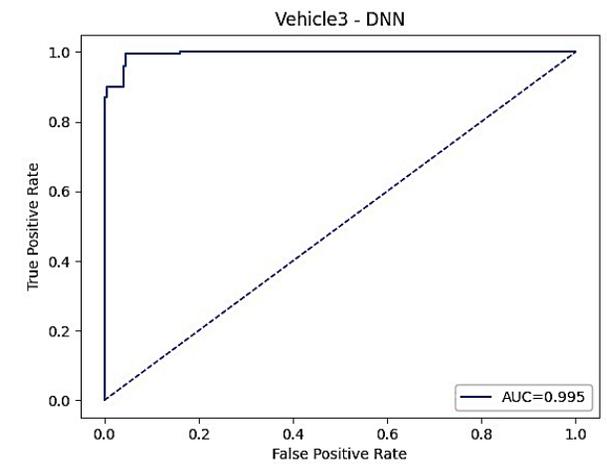

(u)

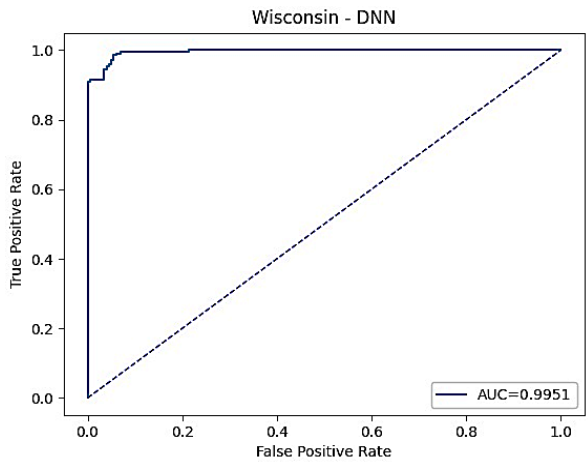 (v)
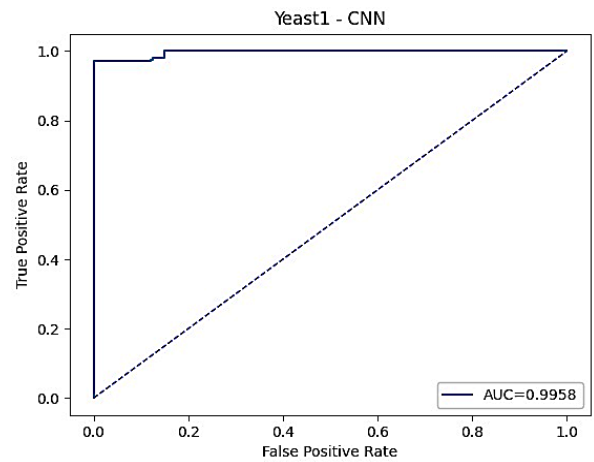 (w)
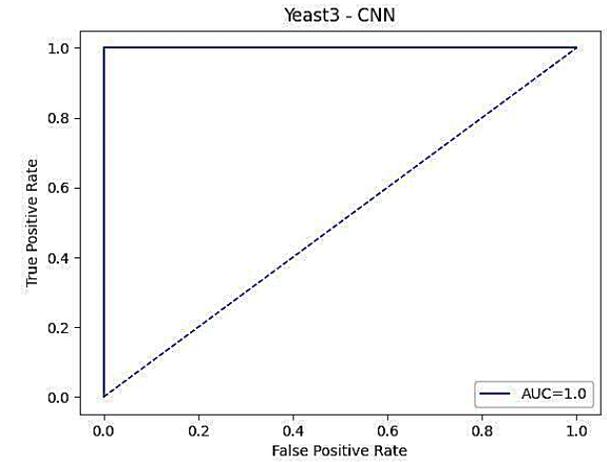 (x)
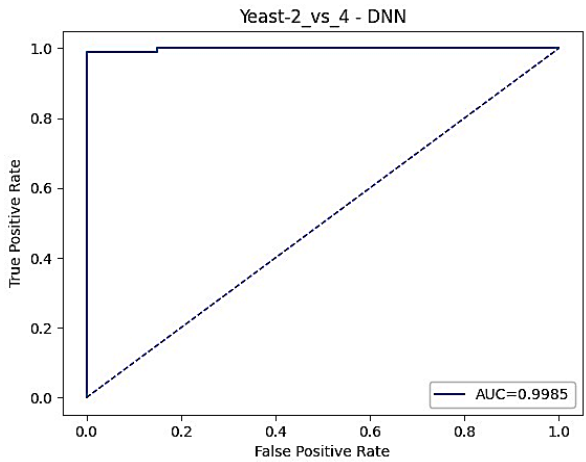 (y)
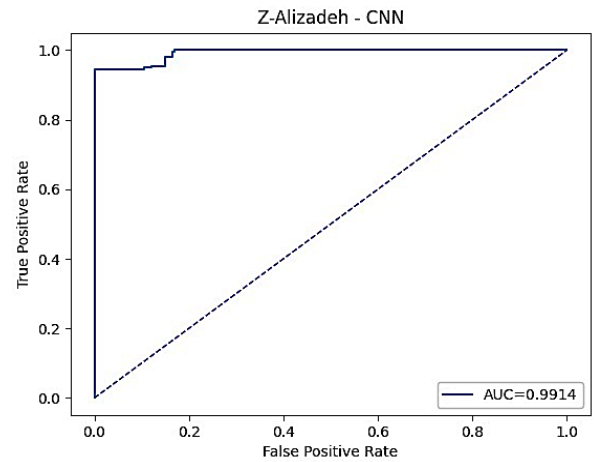 (z)

Fig. 18. ROC plots for the models from (a)-(z).

## 5. Discussion

The classification performance usually drops and faces different failures in the presence of imbalanced datasets. However, imbalanced datasets exist in a broad range of real-life research. Hence, for imbalanced binary classification problems, samples are usually categorized into two classes, majority and minority. Generally, the minority class often illustrates the more significant and crucial samples and interests rather than the majority class samples. Nevertheless, compared to the minority class, the majority class has a larger amount of samples, and in some cases, the situation may be exceedingly serious. Therefore, handling these problems efficiently has become a crucial and significant topic in machine and deep learning methods.

To overcome these challenges, we proposed two methods that are based on DNN and CNN algorithms. At first, several classical and well-known undersampling and oversampling methods such as RUS, Tomek Links, OSS, Near Miss, ROS, and SMOTE were used in the data preprocessing procedure. Also, to achieve better performance, we normalized current datasets. Then, we considered the focal loss function in the process of training the desired models, which are widely implemented in neural networks frameworks for class imbalance problems. Moreover, due to the limited amount of datasets samples which causes unstable classification results, we have trained and evaluated our models for 100 runs and 2000 epochs. In the end, we analyzed our proposed models concerning the accuracy, precision, recall, F1-score, G-Mean, specificity, AUC, and Kappa as evaluation metrics. Based on 24 imbalanced datasets, the average performance score of the evaluation metrics of the executed models is indicated in Table 11.

Table 11. The average performance score of the evaluation metrics of the executed models.

| Models | Acc | Pre | Sen | F1 | G-Mean | Spe | AUC | Kap |
|---|---|---|---|---|---|---|---|---|
| TL + NORM. + CNN | 98.92 | 98.93 | 98.92 | 98.92 | 98.92 | 98.90 | 99.07 | 98.87 |
| TL + NORM. + DNN | 98.95 | 98.90 | 98.90 | 98.90 | 98.90 | 98.87 | 98.98 | 98.79 |
| OSS + NORM. + CNN | 98.78 | 98.79 | 98.78 | 98.78 | 98.78 | 98.72 | 98.93 | 98.73 |
| OSS + NORM. + DNN | 98.82 | 98.83 | 98.82 | 98.82 | 98.82 | 98.80 | 98.94 | 98.72 |
| NearMiss + NORM. + CNN | 98.98 | 98.98 | 98.98 | 98.98 | 98.98 | 98.92 | 99.07 | 98.89 |
| NearMiss + NORM. + DNN | 99.01 | 99.02 | 99.01 | 99.01 | 99.01 | 98.95 | 98.99 | 98.89 |
| RUS + NORM. + CNN | 98.33 | 98.34 | 98.33 | 98.33 | 98.33 | 98.23 | 98.75 | 98.26 |
| RUS + NORM. + DNN | 98.22 | 98.23 | 98.22 | 98.22 | 98.22 | 98.15 | 98.59 | 98.14 |
| **SMOTE + NORM. + CNN** | **99.08** | **99.09** | **99.08** | **99.09** | **99.08** | **99.03** | **99.08** | **98.92** |
| SMOTE + NORM. + DNN | 98.96 | 99.00 | 98.97 | 98.98 | 98.98 | 98.99 | 99.02 | 98.78 |

*Bold specifies that SMOTE + NORM. + CNN is the most robust model.

According to Table 11, it can be founded that the mixed SMOTE+NORM+CNN model has the best performance with 99.08% accuracy, 99.09% precision, 99.08% sensitivity, 99.09% F1-score, 99.08% G-Mean, 99.03% specificity, 99.08% AUC, and 98.92% kappa. Also, the comparison of performance metrics between our study and study (39) on the same imbalanced dataset is demonstrated in Table 12.

Table 12. The comparison of performance metrics on the same dataset.

| Study | Method | Rec(%) | G-mean(%) | F1(%) |
|---|---|---|---|---|
| Mayabadi and Saadatfar, (39) | DB_HS+SVM | 95.80 | 88.30 | 81.40 |
| | DB_HS+RF | 98.10 | 92.00 | 83.80 |
| | DB_US+SVM | 92.70 | 88.50 | 81.50 |
| | DB_US+RF | 95.60 | 93.80 | 87.90 |
| Current Study | **SMOTE+NORM+CNN** | **99.00** | **99.00** | **98.98** |

According to Table 12, the results show the efficiency of our proposed model performance on 16 imbalanced that 99.00% recall, 99.00% G-Mean, and 98.98% F1-score were attained by the SMOTE+NORM+CNN model. In addition, the results of the proposed SMOTE+NORM.+CNN model is compared with the related works on the Z-Alizadeh Sani dataset, as represented in Table 13.

Table 13. The comparison of the metrics results between the proposed study and related studies on the Z-Alizadeh Sani dataset.

| Study | Method | Acc(%) | Rec(%) | F1(%) | Pre(%) | Spe(%) | AUC(%) |
|---|---|---|---|---|---|---|---|
| Alizadehsani et al. 2012, (82) | Sequential Minimal Optimization | 92.09 | 97.22 | NC | NC | 79.31 | NC |
| Alizadehsani et al. 2012, (83) | Ensemble of Naïve Bayes and Sequential Minimal Optimization | 88.52 | 91.12 | NC | NC | 82.05 | NC |
| Alizadehsani et al. 2013, (84) | Information gain+Sequential Minimal Optimization | 94.08 | 96.30 | NC | NC | 88.51 | NC |
| Babič et al., 2017 (85) | Suppoort vector machine | 86.67 | NC | NC | NC | NC | NC |
| Arabasadi et al., 2017, (86) | Neural network+ Genetic algorithm | 93.85 | 97.00 | NC | NC | 92.00 | NC |
| Li et al., 2018, (87) | Naïve Bayes+ Genetic algorithm | 88.16 | 88.00 | NC | NC | 87.78 | NC |

| Study | Method | | | | | | |
|---|---|---|---|---|---|---|---|
| Abdar et al., 2019, (88) | nested ensemble nu-Support Vector Classification+ genetic search algorithm+ multi-step data balancing | 94.66 | 94.70 | 94.70 | 94.70 | NC | 96.60 |
| Abdar et al., 2019, (89) | N2Genetic optimizer-nuSupport Vector Machine | 93.08 | NC | 91.51 | NC | NC | NC |
| Khan et al., 2019, (90) | Neural network + Gini Index for feature selection+ Backward Weight Optimization | 88.49 | NC | NC | NC | NC | NC |
| Kolukısa et al., 2019, (91) | Ensemble Classifier with Fisher Linear Discriminant Analysis | 92.07 | 94.00 | 94.40 | NC | 87.40 | 95.30 |
| Nasarian et al., 2020, (92) | Heterogeneous hybrid feature selection algorithm + SMOTE + Extreme gradient boosting | 92.58 | 92.99 | 90.62 | 92.59 | NC | NC |
| Shahid and Singh, 2020, (93) | Hybrid Particle Swarm Optimization-emotional neural networks coupled with feature selection | 88.34 | 91.85 | 92.12 | 92.37 | 78.98 | NC |
| Ghiasi et al., 2020, (94) | Classification and Regression tree | 92.41 | 98.61 | NC | NC | 77.01 | NC |
| Joloudari et al., 2020, (95) | Random trees | 91.47 | NC | NC | NC | NC | 96.70 |

| Zomorodi-moghadam et al., 2021, (96) | Hybrid Particle Swarm Optimization | 84.25 | NC | NC | NC | NC | NC |
|---|---|---|---|---|---|---|---|
| Ashish et al., 2021, (97) | Support Vector Machine - Extreme gradient boosting+Random forest | 93.86 | NC | 91.86 | 93.86 | NC | NC |
| Zhang et al., 2022, (81) | Extreme gradient boosting + Feature construction +SMOTE | 94.70 | 96.10 | 94.60 | 93.40 | 93.20 | 98.00 |
| Gupta et al., 2022, (98) | Fixed analysis of mixed data+ Binary Bat Algorithm+ Ensemble of Random Forest and Extra Trees | 97.37 | 98.15 | 98.15 | NC | 95.45 | 96.80 |
| **The proposed study** | Mixed SMOTE-NORM.-CNN | 98.57 | 98.58 | 98.57 | 98.58 | 98.42 | 99.14 |

*NC: Not Considered

Based on Table 13, the outcomes show the dominance of the Proposed SMOTE+NORM.+CNN model compared with other studies. The hybrid model of SMOTE+NORM.+CNN verifies the best performance with 98.57% accuracy, 98.58% recall, 98.57% F1-score, 98.58% precision, 98.42% specificity, and 99.14% AUC. In addition to what I have just explained, I can add that the F1-score metric is reputable with the imbalanced Z-Alizadeh Sani dataset because this metric shows the balance between recall and precision for classifiers.

**6. Conclusion and future work**

An unbalanced dataset of the majority and minority classes is a challenging issue when samples belonging to one or more classes are not evenly distributed. Especially, imbalanced dataset reasons deep learning-based models to obtain biased results for binary classification. To address this issue, we presented oversampling and undersampling techniques such as SMOTE, TL, OSS, NearMiss, ROS, and RUS. Among these techniques, SMOTE is the most common robust that targets the growth of the amount of minority class samples by generating synthetic samples, which is employed for balancing datasets with an extremely unbalanced ratio. In this study, six deep learning-based models were used to classify the majority and minority classes. We investigated SMOTE + NORM. + CNN/DNN, TL + NORM. + CNN/DNN, OSS + NORM. + CNN/DNN, NearMiss + NORM. + CNN/DNN, ROS + NORM. + CNN/DNN, and RUS + NORM. + CNN/DNN. To evaluate these models, we utilized KEEL, breast cancer, and Z-Alizadeh Sani datasets. The results show that the mixed SMOTE-NORM-CNN model significantly outperforms other models achieving 99.08% accuracy, 99.09% precision, 99.08% sensitivity, 99.09% F1-score, 99.08% G-Mean, 99.03% specificity, 99.08% AUC, and 98.92% kappa on 24 imbalanced datasets. Also, the proposed model has been compared to the study (39), and the mixed model is suitable for the same dataset. Furthermore, we investigated the related methodologies on the Z-Alizadeh Sani dataset. The results indicate that our proposed methodology is more robust. In future work, we need to apply the proposed model for other applications such as intrusion detection and fraud detection and use the different deep learning architectures models combined with metaheuristic algorithms on other real datasets.

**Conflict of Interest**
The authors declare that the research was conducted in the absence of any commercial or financial relationships that could be construed as a potential conflict of interest.

**Author Contributions**
JHJ designed the study. JHJ, AM, and MAN performed conceptualization and Methodology. The software and analysis of data have been done by AM. Also, JHJ, AM, MAN, SSO and SH wrote the original draft and visualized the figures. The final manuscript has been reviewed and edited by SSO and SH. SSO arranged the funding. All authors have read and approved the final manuscript.

**Acknowledgments**
Not applicable.

43

43